\documentclass{article} % For LaTeX2e
\usepackage{iclr2024_conference,times}

\usepackage{amsmath,amsfonts,bm}

\def\eqref#1{equation~\ref{#1}}
\def\1{\bm{1}}

\DeclareMathAlphabet{\mathsfit}{\encodingdefault}{\sfdefault}{m}{sl}
\SetMathAlphabet{\mathsfit}{bold}{\encodingdefault}{\sfdefault}{bx}{n}

\usepackage{hyperref}
\usepackage{url}
\usepackage{pifont}
\usepackage{soul} 
\usepackage{color, xcolor}

\usepackage{wrapfig,lipsum,booktabs}

\newcommand{\OursMethod}{\emph{Direct Inversion}\xspace}
\newcommand{\Benchmark}{\emph{PIE-Bench}\xspace}
\usepackage{colortbl}
\definecolor{lightyellow}{RGB}{255,242,204}
\definecolor{lightorange}{RGB}{251,229,214}
\definecolor{lightgreen}{RGB}{226,240,217}
\definecolor{lightblue}{RGB}{222,235,247}
\definecolor{lightgray}{RGB}{209,201,206}
\definecolor{deepgray}{RGB}{178,164,173}
\definecolor{deepblue}{RGB}{112,168,218}

\usepackage{threeparttable}
\usepackage{xspace}
\usepackage{graphicx} 
\usepackage{float} 
\usepackage{subfigure} 
\usepackage{booktabs}
\usepackage{multicol}
\usepackage{multirow} 
\usepackage[ruled,boxed,linesnumbered]{algorithm2e}
\usepackage{caption}

\usepackage{setspace}

\title{Direct Inversion: Boosting Diffusion-based Editing with 3 Lines of Code}

\author{Xuan Ju$^{1,2}$\thanks{This work was done when Xuan Ju was intern at IDEA.}~,~Ailing Zeng$^{2}$\thanks{Corresponding author.}~,~Yuxuan Bian$^{1}$,~Shaoteng Liu$^{1}$,~Qiang Xu$^{1}$$^{\dag}$ \\
$^1$The Chinese University of Hong Kong (CUHK) 
$^2$International Digital Economy Academy (IDEA) \\
\texttt{\small{\{xju22,stliu21,qxu\}@cse.cuhk.edu.hk}}
\texttt{\small{\{zengailing\}@idea.edu.cn}}\\
}

\iclrfinalcopy % Uncomment for camera-ready version, but NOT for submission.
\begin{document}

{%
\maketitle
\begin{figure}[H]
\centering
\vspace{-1.2cm}
\includegraphics[width=0.98\textwidth]{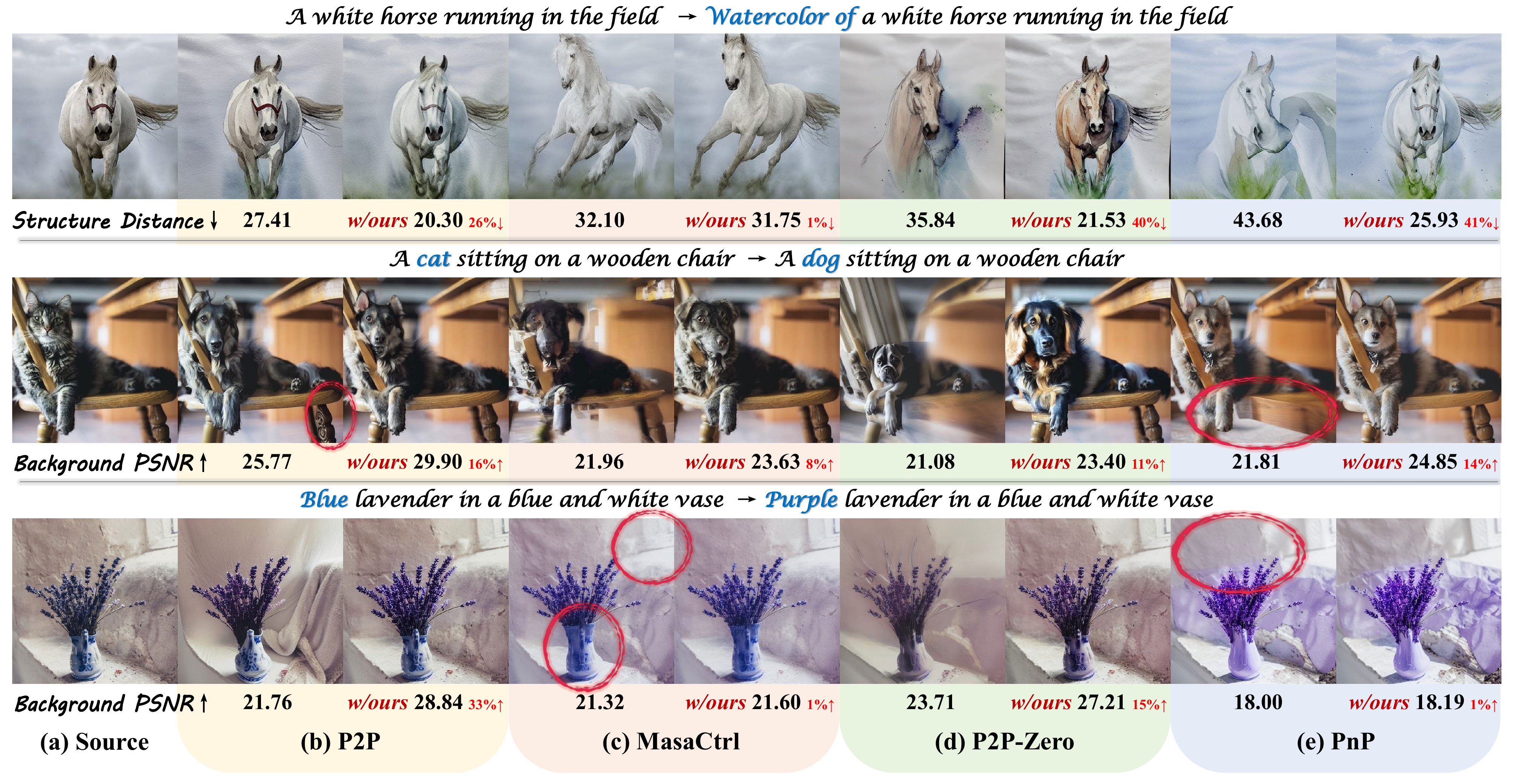}
\vspace{-0.3cm}
\captionsetup{font={stretch=0.8}}
\caption{\textbf{Performance enhancement of incorporating \OursMethod into four diffusion-based editing methods} across various editing categories (from top to bottom): style transfer, object replacement, and color change. The editing prompt is displayed at the top of each row, which includes (a) the source image, the editing results of \sethlcolor{lightyellow}\hl{(b) Prompt-to-Prompt (P2P)}~\citep{hertz2022prompt}, \sethlcolor{lightorange}\hl{(c) MasaCtrl}~\citep{cao2023masactrl}, \sethlcolor{lightgreen}\hl{(d) pix2pix-zero}~\citep{parmar2023zero}, and \sethlcolor{lightblue}\hl{(e) plug-and-play}~\citep{tumanyan2023plug}. Each set of results is presented: the first column w/o \OursMethod (Null-text inversion for P2P, DDIM Inversion for the others), and the second column w/ \OursMethod. Incorporating \OursMethod into diffusion-based editing methods results in improved image structure preservation (enhancement of the structure distance metric) for full image editing and enhanced background preservation (increased PSNR metric values in the background, i.e., areas that should remain unedited) for foreground editing. The improvements are mostly tangible, and we circle some of the subtle discrepancies w/o \OursMethod in red. %Furthermore, the integration of \OursMethod maintains the editing capabilities of existing editing methods. 
\textbf{Best viewed with zoom in.}}
\vspace{-0.5cm}
\label{fig:teaser}
\end{figure}
}

\begin{abstract}
\vspace{-0.4cm}
\begin{spacing}{0.9}
Text-guided diffusion models have revolutionized image generation and editing, offering exceptional realism and diversity. Specifically, in the context of diffusion-based editing, where a source image is edited according to a target prompt, the process commences by acquiring a noisy latent vector corresponding to the source image via the diffusion model. This vector is subsequently fed into separate source and target diffusion branches for editing. The accuracy of this inversion process significantly impacts the final editing outcome, influencing both \emph{essential content preservation} of the source image and \emph{edit fidelity} according to the target prompt.

Prior inversion techniques aimed at finding a unified solution in both the source and target diffusion branches. However, our theoretical and empirical analyses reveal that disentangling these branches leads to a distinct separation of responsibilities for preserving essential content and ensuring edit fidelity. Building on this insight, we introduce ``\OursMethod,'' a novel technique achieving optimal performance of both branches with just three lines of code. To assess image editing performance, we present \Benchmark, an editing benchmark with $700$ images showcasing diverse scenes and editing types, accompanied by versatile annotations and comprehensive evaluation metrics. Compared to state-of-the-art optimization-based inversion techniques, our solution not only yields superior performance across $8$ editing methods but also achieves nearly an order of speed-up. 
\end{spacing}
\end{abstract}

\section{Introduction}\label{sec:Introduction}

\begin{figure}[ht]
\vspace{-0.2cm}
        \begin{center}
            \includegraphics[width=1.0\textwidth]{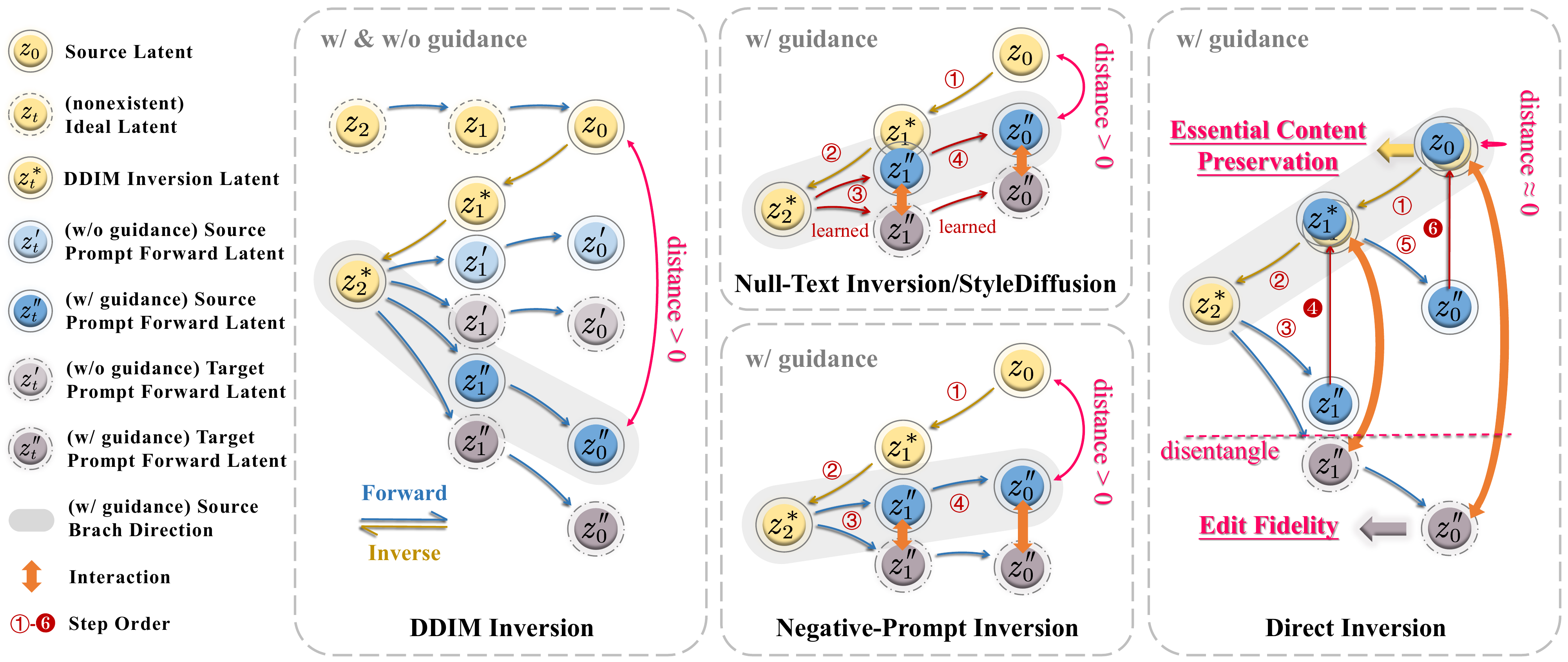}
        \end{center}
\vspace{-0.3cm}
    \caption{\textbf{Comparisons among different inversion methods in diffusion-based editing.} We assume a 2-step diffusion process for illustration. Due to nonexistent of ideal \sethlcolor{lightyellow}\hl{$z_2$}, common practice uses DDIM Inversion~\citep{song2020denoising} to approximate \sethlcolor{lightyellow}\hl{$z_t$}, resulting in \sethlcolor{lightyellow}\hl{$z_{t}^{*}$}  with perturbation. Diffusion-based editing methods start from the perturbed noisy latent \sethlcolor{lightyellow}\hl{$z_{2}^{*}$} and perform DDIM sampling in a \sethlcolor{lightblue}\hl{sou}\sethlcolor{deepblue}\hl{rce} and a \sethlcolor{lightgray}\hl{tar}\sethlcolor{deepgray}\hl{get} diffusion branch, further resulting in the \textcolor[RGB]{255,1,98}{distance} shown on the figure. Null-Text Inversion and StyleDiffusion optimize a specific latent used in both source and target branches to reduce this distance. Negative-Prompt Inversion assigns the guidance scale to $1$ to decrease the distance. In contrast, \OursMethod disentangles source and target branches in editing. By leaving the target diffusion branch untouched, \OursMethod retains the edit fidelity. By directly returning the source branch to \sethlcolor{lightyellow}\hl{$z_0$}, \OursMethod achieves the best possible essential content preservation.
    }
        \label{fig:inverse}
\end{figure}

Text-guided diffusion models~\citep{rombach2022high,ramesh2022hierarchical} have become the mainstream image generation technique, praised for their realism and diversity.
As the noise latent space of diffusion models~\citep{meng2022sdedit,kawar2023imagic,hertz2022prompt,balaji2022ediffi,liu2023video} possesses the capacity to retain and modify images, we can perform prompt-based editing with diffusion models, where a source image is edited according to a target prompt.
The common practice is to maintain two diffusion branches: one for the source image and the other for the target image. By carefully exchanging information between the two branches, we can preserve the essential content in the source image while achieving edit fidelity according to the target prompt.
However, such manipulations are only straightforward when the diffusion latent space (noisy latent in each diffusion step) corresponding to the source image is available. When editing images without known latent space, we have to invert the diffusion model to obtain their latent vectors first.

While DDIM inversion is effective for unconditional diffusion~\citep{song2020denoising,dhariwal2021diffusion}, much of the research~\citep{hertz2022prompt,han2023improving} has centered on inverting the diffusion process with conditional inputs. This is driven by the significance of conditions in applications like text-based image editing.
However, introducing conditions undermines DDIM inversion quality, as evidenced in Figure~\ref{fig:inverse}.
With the advent of Null-Text Inversion~\citep{mokady2023null}, a prevailing consensus~\citep{dong2023prompt,li2023stylediffusion} has emerged: achieving superior inversion\footnote{A more suitable name for them would be inversion correction techniques. But we follow previous works and call them inversion techniques.} necessitates rigorous optimization. Methods that forgo such optimization, such as Negative-Prompt Inversion~\citep{miyake2023negative}, compromise editing outcomes.  
In this paper, we challenge this prevailing wisdom, posing two fundamental questions: \textbf{What exactly are these optimization-based inversion methods truly aiming at? And, are such optimizations indispensable for diffusion-based image editing?}

As illustrated in Figure \ref{fig:inverse}, prior optimization-based approaches strive to minimize the distance between $z_0$ and $z_{0}^{''}$ by indirectly tweaking the generation model's input parameters. Given the magnitude of the optimization network, like UNet, and the impracticality of prolonged optimization durations, these methods often restrict the optimization of the target latent to just a few iterations. This results in a learned latent with a discernible gap between $z_{0}^{''}$ and the original $z_0$. Moreover, the learned latent does not appear during the generation models' training process, deviating from the pretrained diffusion model's original input distribution. Such forced input assignments hinder the model's generative capacity, compromising the integrity of both the source and editing branches.

In this work, we delve into the intricacies of text-based inversion, providing a thorough analysis of existing techniques. Our theoretical and empirical analyses demonstrate that the exhaustive optimization in prior techniques is, counterintuitively, not only far from optimal but in fact unnecessary. Introducing \textbf{\OursMethod}, our approach offers a simple yet potent inversion solution for diffusion-based editing. The essence of \OursMethod lies in two primary strategies: (1) \textbf{disentangle} the source and target branches, and (2) empower each branch to \textbf{excel in its designated role}: preservation or editing. Specifically, the source branch in \OursMethod rectifies the deviation path directly, using only $3$ lines of code. This addresses the challenges seen in earlier approaches: (1) undesirable latent space distances affecting \textbf{essential content preservation}, (2) misalignment in the generation model's distribution, and (3) extended processing times. For the target branch, we keep it unchanged, ensuring the best possible \textbf{edit fidelity} in line with the target prompt.

Addressing the lack of standardized benchmarks for inversion and editing, we introduce \textbf{PIE-Bench} (\textbf{P}rompt-based \textbf{I}mage \textbf{E}diting \textbf{Bench}mark) with $700$ images from diverse scenes, spanning $10$ unique editing categories. Each entry consists of a source prompt, target prompt, editing directive, edit subjects, and a hand-annotated editing mask. To rigorously assess \OursMethod and benchmark it against existing techniques, we employ $7$ metrics emphasizing both essential content preservation and edit fidelity. 
Compared with $5$ inversion methods with Prompt-to-Prompt editing, \OursMethod outperforms them, enhancing essential content preservation by up to $83.2\%$ in Structure Distance, up to $73.9\%$ in background LPIPS, and edit fidelity by up to $8.8\%$ in Edit Region Clip Similarity, while achieving nearly an order of editing speedup over optimization-based inversion methods. Moreover, across $8$ editing approaches, \OursMethod boosts content preservation by as much as $20.2\%$ and edit fidelity by up to $2.5\%$ relative to their baseline configurations. Visualization results are shown in Figure~\ref{fig:teaser}. \footnote{Code is available at  \url{https://github.com/cure-lab/DirectInversion}.}

\vspace{-0.3cm}

\section{Related Work}\label{sec:Related_Work}

\vspace{-0.2cm}

\begin{figure}
        \begin{center}
            \includegraphics[width=0.95\textwidth]{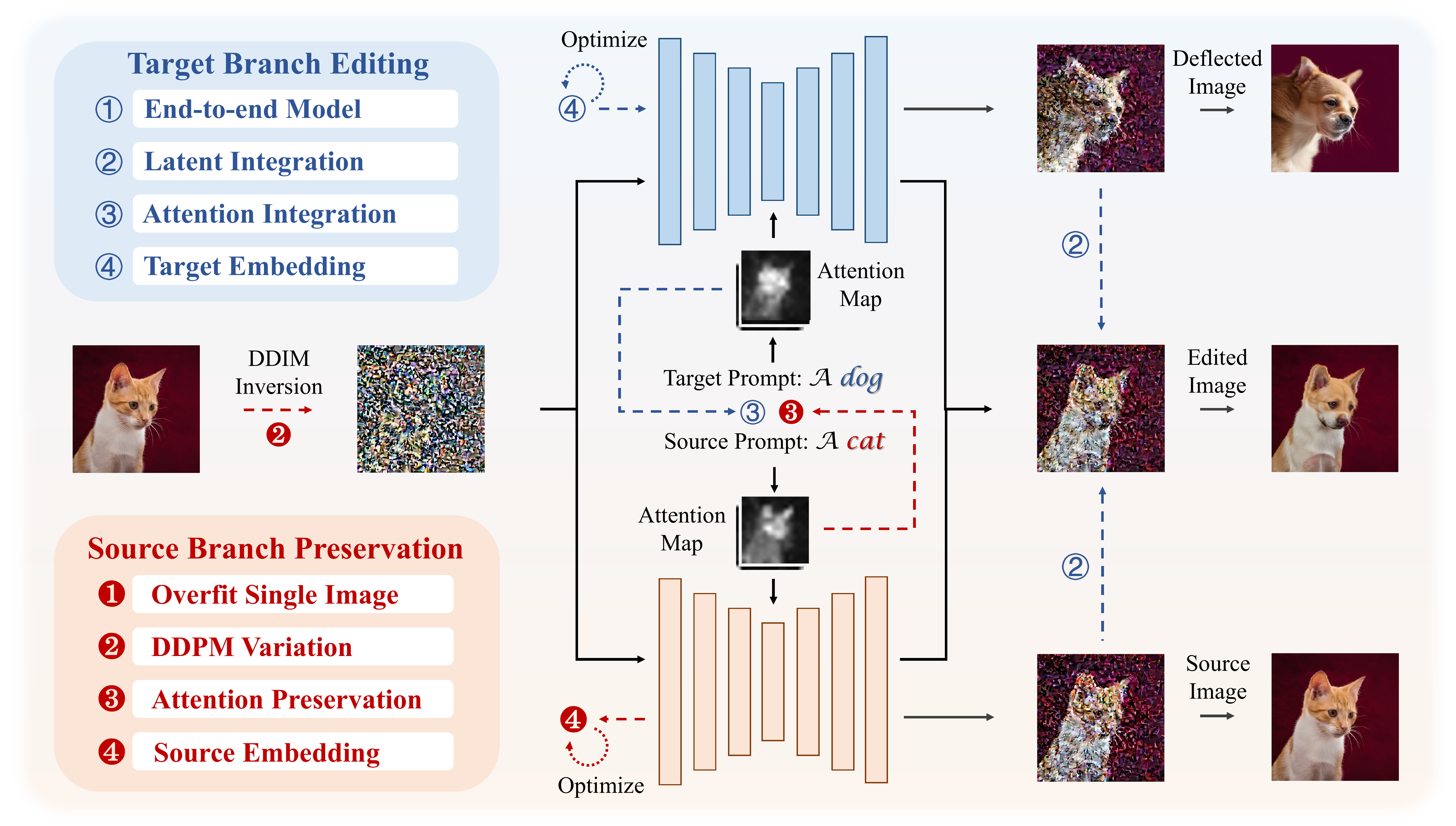}
        \end{center}
        \vspace{-0.3cm}
    \caption{\textbf{Diffusion-based editing pipline} showing how edit fidelity and essential content preservation are achieved in different methods. Detailed introduction is in supplementary files.}
        \label{fig:related_work}
    \vspace{-0.5cm}
\end{figure}

Diffusion-based image editing aims to manipulate images with diffusion models using given instructions such as natural language descriptions~\citep{hertz2022prompt}, point dragging~\citep{shi2023dragdiffusion}, and semantic masking~\citep{meng2022sdedit}.
This involves two primary concerns: 
(1) edit fidelity: ensure the editing aligns with the provided instructions; and 
(2) essential content preservation: inverse the images, particularly the regions that do not require modification, into diffusion latent space while ensuring accurate reconstruction during the editing procedure. 
Accordingly, we undertake a comprehensive review of prior methodologies concerning both two aspects, as shown in Figure~\ref{fig:related_work}.

For edit fidelity, previous methods perform editing roughly through four ways: 
\ding{172} end-to-end editing model~\citep{brooks2023instructpix2pix, kim2022diffusionclip, nichol2021glide,geng2023instructdiffusion} that trains end-to-end diffusion models to edit images
, \ding{173} latent integration~\citep{meng2022sdedit, avrahami2022blended, avrahami2023blended, couairon2022diffedit, zhang2023sine, shi2023dragdiffusion,joseph2023iterative} that inserts editing instruction through the noisy latent feature in target diffusion branch to source diffusion branch.
, \ding{174} attention integration~\citep{hertz2022prompt, han2023improving, parmar2023zero, cao2023masactrl, tumanyan2023plug, zhang2023real, mou2023dragondiffusion} that fuses attention map connecting the text and image in the source and editing diffusion branch, and \ding{175} target embedding~\citep{kawar2023imagic, cheng2023general, wu2023uncovering, brack2023sega, tsaban2023ledits, valevski2022unitune, dong2023prompt} that aggregates editing information of the target branch into an embedding and then insert it to source diffusion branch.

For essential content preservation, previous methods tried to retain the source image's feature through \ding{182} overfit the editing image~\citep{kawar2023imagic,shi2023dragdiffusion} so that editing will not make massive changes to the image content, \ding{183} DDPM/DDIM inversion variation~\citep{miyake2023negative,huberman2023edit,wallace2023edict} to strengthen the source image's influence on both the source and target branch by modifying DDPM/DDIM inversion formulation, \ding{184} attention preservation~\citep{mou2023dragondiffusion,cheng2023general,cao2023masactrl,parmar2023zero,tumanyan2023plug,hertz2022prompt,qi2023fatezero} that retains the attention map feature of the source diffusion branch during attention map fusion of the source and target branches, and \ding{185} source embedding~\citep{mokady2023null, dong2023prompt, li2023stylediffusion,gal2022image,fei2023gradient,huang2023reversion} that absorbs the background or image information to an embedding and use this embedding to reconstruct essential content of the source image.

Due to the page limit, we put the detailed introduction of related work in supplementary files.

\vspace{-0.3cm}

\section{Preliminaries}\label{sec:Preliminaries}

\vspace{-0.2cm}

This section will first introduce DDIM sampling (DDIM Forward and DDIM Inverse) and the classifier-free guidance commonly employed in Diffusion Models. Then, we will delve into the issues arising from the utilization of classifier-free guidance and DDIM sampling in editing, and show how previous related works, including Null-Text Inversion~\citep{mokady2023null}, StyleDiffusion~\citep{li2023stylediffusion}, and Negative-prompt Inversion~\citep{miyake2023negative} address these challenges.

\vspace{-0.2cm}

\paragraph{Diffusion Models.} 

Diffusion models aim to map a random noise vector $z_T$ to a series of noise samples $z_{t}$ and, finally, an output image or latent $z_{0}$ by adding Gaussian noise $\epsilon$ step-by-step, where $t\sim \left[ 1,T \right]$ and $T$ is the timestep number. To train diffusion models, we first sample $z_{t}$ from a real image following \eqref{eq:add_noise} where $\epsilon \sim \mathcal{N} \left( 0,1 \right)$ and $\alpha$ is hyper-parameter.

\vspace{-0.2cm}

\begin{equation}
z_t=\sqrt{\alpha _t}z_{0}+\sqrt{1-\alpha _t}\epsilon 
 \label{eq:add_noise}
\end{equation}

\vspace{-0.2cm}

Then, a denoiser network $\epsilon  _{\theta}$ is optimized to predict $\epsilon(z_t, t)$ with the objective:

\vspace{-0.2cm}

\begin{equation}
\underset{\theta}{\min}E_{z_0,\epsilon \sim \mathcal{N} \left( 0,I \right) ,t\sim Uniform\left( 1,T \right)}\left\| \epsilon -\epsilon _{\theta}\left(z_t, t\right) \right\| 
 \label{eq:train_objective}
\end{equation}

\vspace{-0.2cm}

To generate images from given $z_T$, we employ the deterministic DDIM sampling~\citep{song2020denoising}:

\vspace{-0.2cm}

\begin{equation}
z_{t-1}=\frac{\sqrt{\alpha _{t-1}}}{\sqrt{\alpha _t}}z_t+\sqrt{\alpha _{t-1}}\left( \sqrt{\frac{1}{\alpha _{t-1}}-1}-\sqrt{\frac{1}{\alpha _t}-1} \right) \epsilon _{\theta}\left( z_t,t \right) 
 \label{eq:ddim_sample}
\end{equation}

\vspace{-0.2cm}

\paragraph{DDIM Inversion.} 
Although diffusion models have superior characteristics in the feature space~\citep{balaji2022ediffi,dong2023prompt,feng2022training} that can support various downstream tasks, similar to GAN-based models~\citep{xia2022gan}, it is hard to apply them to images in the absence of natural diffusion feature space for non-generated images. Thus, a technique inverting $z_{0}$ back to $z_{T}$ is necessary. To address this, a straightforward inversion technique known as DDIM inversion is commonly used for unconditional diffusion models, predicated on the presumption that the ODE process can be reversed in the limit of infinitesimally small steps.

\vspace{-0.2cm}

\begin{equation}
z_{t}^{*}=\frac{\sqrt{\alpha _t}}{\sqrt{\alpha _{t-1}}}z_{t-1}^{*}+\sqrt{\alpha _t}\left( \sqrt{\frac{1}{\alpha _t}-1}-\sqrt{\frac{1}{\alpha _{t-1}}-1} \right) \epsilon _{\theta}\left( z_{t-1}^{*},t-1 \right) 
 \label{eq:ddim_inversion}
\end{equation}

\vspace{-0.2cm}

However, in most diffusion models, this presumption cannot be guaranteed, resulting in a perturbation from $z_t$ to $z_t^*$ in \eqref{eq:ddim_sample}, \eqref{eq:ddim_inversion} and Figure \ref{fig:inverse}. Consequently, an additional perturbation from $z_{t}^{*}$ to $z_{t}^{\prime}$ arises when utilizing \eqref{eq:add_noise} to sample an image from $z_{T}^{*}$ as shown in Figure~\ref{fig:inverse}.

\vspace{-0.2cm}

\paragraph{Classifier-free Guidance.} 

Previously, we only considered scenarios without any associated conditions. To insert conditions such as text, Ho et al.~\citep{ho2022classifier} present the classifier-free guidance, where the prediction is performed both unconditionally and conditionally, and then mixed together through:

\vspace{-0.2cm}

\begin{equation}
\epsilon _{\theta}\left( z_t,t,C,\oslash \right) =w\cdot \epsilon _{\theta}\left( z_t,t,C \right) +\left( 1-w \right) \cdot \epsilon _{\theta}\left( z_t,t,\oslash \right), 
 \label{eq:classifier_free_guidance}
\end{equation}

\vspace{-0.2cm}

where $w$ is the guidance scale, $C$ is the condition (embedding of text prompt in our task), and $\oslash$ is the null condition (embedding of null in our task). This further leads to another perturbation from $z_{t}^{\prime}$ to $z_{t}^{''}$ due to the destruction of the DDIM process as demonstrated in Figure~\ref{fig:inverse}.

\vspace{-0.2cm}

\paragraph{Previous Inversion Techniques.} 

Currently, the predominant inversion technique employed for reducing the adverse impact caused by DDIM inversion and classifier-free guidance is optimization-based methods, such as Null-Text Inversion~\citep{mokady2023null} and StyleDiffusion~\citep{li2023stylediffusion}. Alternative inversion techniques, such as Edit Friendly DDPM~\citep{huberman2023edit}, Negative-Prompt Inversion~\citep{miyake2023negative}, and EDICT~\citep{wallace2023edict}, exhibit instability in editing outcomes in both essential content preservation and edit fidelity. The qualitative and quantitative results in our experiment further corroborate this instability.

Optimization-based inversion methods use a specific latent variable to reduce the difference between $z_{t}^{''}$ and $z_{t}^{*}$. For example, Null-Text Inversion revises \eqref{eq:classifier_free_guidance} to \eqref{eq:nul_text_classifier_free_guidance} and learns the specific latent \textcolor{red}{variable} by gradient propagation using the loss function $z_{t}^{''} - z_{t}^{*}$. This learned \textcolor{red}{variable} will be further used in both the source and target branches in editing.

\vspace{-0.2cm}

\begin{equation}
\epsilon _{\theta}\left( z_t,t,C,\oslash \right) =w\cdot \epsilon _{\theta}\left( z_t,t,C \right) +\left( 1-w \right) \cdot \epsilon _{\theta}\left( z_t,t,{\color[RGB]{240, 0, 0}variable} \right) 
 \label{eq:nul_text_classifier_free_guidance}
\end{equation}

\vspace{-0.2cm}

\vspace{-0.2cm}

\section{Method}\label{sec:Method}

\vspace{-0.2cm}

\subsection{Motivation}\label{sec:Motivation}

\vspace{-0.2cm}

We explain our motivation by raising and then answering two questions.

\textbf{Why do optimization-based methods perform better among previous inversion methods?}

Edit Friendly DDPM~\citep{huberman2023edit} proposes an alternative latent noise space by changing the DDPM sampling distribution to help reconstruction of the desired image.
Negative-Prompt Inversion~\citep{miyake2023negative} assigns conditioned text embedding to Null-Text embedding and thus maintains a guidance scale of $1$ to reduce the deviation in editing.
EDICT~\citep{wallace2023edict} maintains two coupled noise vectors to invert each other in an alternating fashion for image reconstruction, which reduces the editability of diffusion models. 
Compared with these inversion techniques, optimization-based inversion~\citep{mokady2023null,li2023stylediffusion,dong2023prompt} does not influence the distribution in DDIM sampling (compared to Edit Friendly DDPM), retains enough guidance for text conditions (compared to Negative-Prompt Inversion), and maintains the diffusion model's editability (compared to EDICT).

\textbf{Are such optimizations indispensable and optimal for diffusion-based image editing?}

Optimization-based inversion methods learn a specific latent variable to minimize the loss function $z_{t}^{''} - z_{t}^{*}$. Thus, the target of optimization-based inversion is to correct $z_{t}^{''}$ back to $z_{t}^{*}$. The learned latent variable then serves as an input for both the source and target branches.

The optimization of a unified variable for source and target branches leads to several problems: (1) To optimize the specific latent variable, a prolonged processing time is needed during inference (\emph{e.g.}, 148.48 seconds per image for Null-Text Inversion), which is impractical for editing tasks with user interaction; (2) Considering that extended optimization times are not expected, previous approaches have opted to optimize the target latent for only a limited number of iterations. Consequently, the result frequently entails a learned latent space with a discernible gap between $z_{0}^{''}$ and the initial $z_0$, especially when a large distance exists between $z_{t}^{''}$ and $z_{t}^{*}$. This leads to a decline in essential content preservation ability, as shown in our ablation study; (3) The learned variable serves as the generation model's input parameter, which is not aligned with the diffusion model's expected input distribution and leads to negative impacts on the diffusion model integrity. These three issues hinder the practicality and editability of these optimization-based inversion methods.

\vspace{-0.2cm}

\subsection{Method}

\vspace{-0.2cm}

Bearing these issues into consideration, we propose \OursMethod. The \textbf{key} of \OursMethod is to \textbf{disentangle the source and target branch}, thus enabling each branch to unleash its maximum potential individually. In the source branch, we can directly add $z_{t}^{*}-z_{t}^{''}$ back to $z_{t}^{''}$, which is a simple strategy that can directly rectify the deviation path and is plug-and-play to various editing methods. In the target branch, simply leaving it unaltered would maximize the diffusion models' potential for target image generation. This simple but effective solution solves the three issues in optimized-based inversion by (1) No optimization is required, thus incurring minimal additional time overhead; (2) Adding $z_{t}^{*}-z_{t}^{''}$ eliminates the discernible gap between $z_{0}^{''}$ and the initial $z_0$; (3) Do not have any impact on the distribution of the diffusion model's input.

\begin{algorithm}[h]
\small
\SetAlgoLined
\SetKwInOut{KwResult}{Part \uppercase\expandafter{\romannumeral1}}
\SetKwInOut{KwData}{Part \uppercase\expandafter{\romannumeral2}}
\KwIn{A source prompt embedding $C^{src}$ (or embedding of null for some editing methods), a target prompt embedding $C^{tgt}$, a real image or latent embedding $z_{0}^{src}$}
\KwOut{An edited image or latent embedding $z_{0}^{tgt}$}
 \vspace{1mm} \hrule \vspace{1mm}
 \KwResult{\textbf{Inverse $z_{0}^{src}$}}
 \vspace{1mm} \hrule \vspace{1mm}
 $z_{0}^{*}=z_{0}^{src}$;\\
 \For{$t=1,\ldots,T-1,T$}{
    
    $z_{t}^{*} \leftarrow \mathrm{DDIM}\_\mathrm{Inversion}\left(z_{t-1}^{*},t-1,\left[C^{src},C^{tgt}\right]\right)$;

 }
 \vspace{1mm} \hrule \vspace{1mm}
 \KwData{\textbf{Perform editing on $z_{T}^{*}$ \textcolor[RGB]{232, 0, 0}{with \OursMethod}}}
 \vspace{1mm} \hrule \vspace{1mm}
 $z_{T}^{tgt}=z_{T}^{*}$; $z_{T}^{''}=z_{T}^{*};$
 
 \For{$t=T,T-1,\ldots,1$}{
    \sethlcolor{lightgray}\hl{${\color[RGB]{232, 0, 0}\left[\boldsymbol{o}_{\boldsymbol{t}-\boldsymbol{1}}^{\boldsymbol{src}},\boldsymbol{o}_{\boldsymbol{t}-\boldsymbol{1}}^{\boldsymbol{tgt}}\right] \leftarrow \boldsymbol{z}_{\boldsymbol{t}-\boldsymbol{1}}^{*} - \mathrm{\boldsymbol{DDIM}}\_\mathrm{\boldsymbol{Forward}}\left(\boldsymbol{z}_{\boldsymbol{t}}^{''},\boldsymbol{t},\left[\boldsymbol{C}^{\boldsymbol{src}},\boldsymbol{C}^{\boldsymbol{tgt}}\right]\right)}$} \tcp*{1 {\tiny calculate distance}}
    \sethlcolor{lightgray}\hl{${\color[RGB]{232, 0, 0} \boldsymbol{z}_{\boldsymbol{t}-\boldsymbol{1}}^{''}=\mathrm{\boldsymbol{DDIM}}\_\mathrm{\boldsymbol{Forward}}\left(\boldsymbol{z}_{\boldsymbol{t}}^{''},\boldsymbol{t},\left[\boldsymbol{C}^{\boldsymbol{src}},\boldsymbol{C}^{\boldsymbol{tgt}}\right]\right)+\left[ \boldsymbol{o}_{\boldsymbol{t}-\boldsymbol{1}}^{\boldsymbol{src}},\boldsymbol{0} \right]}$} \tcp*{2 {\tiny update $z_{t-1}^{''}$}}
    $z_{t-1}^{tgt} \leftarrow \mathrm{DDIM}\_\mathrm{Forward} \,_{\mathrm{Editing}\_\mathrm{Model}}\left(z_{t}^{tgt},t,\left[C^{src},C^{tgt}\right]\right)$\sethlcolor{lightgray}\hl{${\color[RGB]{232, 0, 0} + \left[\boldsymbol{o}_{\boldsymbol{t}-\boldsymbol{1}}^{\boldsymbol{src}},\boldsymbol{0}\right]} $} \tcp*{3 \tiny add distance}}
 \textbf{Return} $z_{0}^{tgt}$
 
 \caption{Real Image Editing Pipeline with Direct Inversion}
\label{alg:direct_inversion}
\end{algorithm}

Algorithm~\ref{alg:direct_inversion} presents the algorithm for inserting \OursMethod into existing diffusion-based image editing methods. Red lines with gray backgrounds highlight the $3$ lines of code added by \OursMethod. Typical diffusion-based editing of images involves two parts: an inversion process to get the diffusion space of the image, and a forward process to perform editing on the diffusion space. \OursMethod can be plug-and-played into the forward process and rectifies the deviation path step by step. Specifically, \OursMethod first computes the difference between $z_{t-1}^{*}$ and $z_{t-1}^{''}$, then adds the difference back to $z_{t-1}^{''}$ in DDIM forward. We only add the difference of the source prompt in latent space and update $z_{t-1}^{''}$ with Algorithm~\ref{alg:direct_inversion} line 8 instead of $z_{t-1}^{''}=z_{t-1}^{*}$, which is the keys to retaining the editability of the target prompt's latent space.

\vspace{-0.2cm}

\subsection{Benchmark Construction}

\vspace{-0.2cm}

While diffusion-based editing has garnered significant attention in recent years, evaluations of various editing methods have primarily relied on subjective and limited visualizations. 
To systematically validate our proposed method as a plug-and-play strategy for editing models and compare our method with existing inversion methods, as well as compensate for the absence of standardized performance criteria for inversion and editing techniques, we construct a benchmark dataset, named \Benchmark (\textbf{P}rompt-based \textbf{I}mage \textbf{E}diting \textbf{Bench}mark).

\Benchmark comprises $700$ images featuring $10$ distinct editing types. Images are evenly distributed in natural and artificial scenes (\emph{e.g.}, paintings) among four categories: animal, human, indoor, and outdoor.
Each image in \Benchmark includes five annotations: source image prompt, target image prompt, editing instruction, main editing body, and the editing mask. Notably, the editing mask annotation (indicating the anticipated editing region) is crucial in accurate metrics computations as we expect the editing to only occur within a designated area. Details are in the supplementary files.

\vspace{-0.7cm}

\section{Experiments}\label{sec:Experiments}

\vspace{-0.2cm}

Due to page limitation, we only provide the comparison of inversion-based editing, essential content preservation methods, ablation on \OursMethod and Null-Text Inversion, and the influence of adding the difference to target latent in this section. More experiments are in supplementary files.

\vspace{-0.2cm}

\subsection{Evaluation Metrics}

\vspace{-0.2cm}

To illustrate the effectiveness and efficiency of our proposed \OursMethod, we use eight metrics covering four aspects: structure distance~\citep{tumanyan2022splicing}, background preservation (PSNR, LPIPS~\citep{zhang2018unreasonable}, MSE, and SSIM~\citep{wang2004image} outside the annotated editing mask), edit prompt-image consistency (CLIPSIM~\citep{clipsim} of the whole image and regions in the editing mask) and inference time. Details can be found in the supplementary files.

\vspace{-0.2cm}

\subsection{Comparison with Inversion-based Editing}

\vspace{-0.2cm}

\begin{table}[htbp]
\small
\centering
  \renewcommand\arraystretch{0.8}
\setlength{\tabcolsep}{0.3mm}{
\begin{threeparttable}
\begin{tabular}{c|c|c|cccc|cc}
\toprule
\toprule
\multicolumn{2}{c|}{\textbf{Method}}           & \textbf{Structure}          & \multicolumn{4}{c|}{\textbf{Background Preservation}} & \multicolumn{2}{c}{\textbf{CLIP Similariy}} \\ \midrule
\textbf{Inverse}          & \textbf{Editing}            & \textbf{Distance}$_{^{\times 10^3}}$ $\downarrow$ & \textbf{PSNR} $\uparrow$     & \textbf{LPIPS}$_{^{\times 10^3}}$ $\downarrow$  & \textbf{MSE}$_{^{\times 10^4}}$ $\downarrow$     & \textbf{SSIM}$_{^{\times 10^2}}$ $\uparrow$    & \textbf{Whole}  $\uparrow$          & \textbf{Edited}  $\uparrow$       \\ \midrule
\textbf{DDIM}& \textbf{P2P}                & 69.43           & 17.87  & 208.80  & 219.88  & 71.14 &  \underline{25.01}        &  \textbf{22.44}          \\
\textbf{NT\tnote{\dag}}& \textbf{P2P}                & \underline{13.44}           & \underline{27.03}  & \underline{60.67}  & \underline{35.86}  & \underline{84.11} & 24.75        & 21.86          \\
\textbf{NP}& \textbf{P2P}                &     16.17       &  26.21 &  69.01  &  39.73 & 83.40 &   24.61     &    21.87     \\
\textbf{StyleD}& \textbf{P2P}                &  \textbf{11.65} & 26.05 & 66.10 & 38.63 & 83.42 & 24.78 & 21.72    \\
\midrule
\textbf{Ours}             & \textbf{P2P}                & \textbf{11.65}\textcolor{red}{$_{83\%\downarrow}$}           & \textbf{27.22}\textcolor{red}{$_{52\%\uparrow}$}  & \textbf{54.55}\textcolor{red}{$_{74\%\downarrow}$}  & \textbf{32.86}\textcolor{red}{$_{85\%\downarrow}$}  & \textbf{84.76}\textcolor{red}{$_{19\%\uparrow}$} & \textbf{25.02}$_{\color{red}1.7\%\uparrow}$         & \underline{22.10}$_{{\color{red}1.7\%\uparrow}}$          \\
\midrule
\midrule
\textbf{DDIM}& \textbf{MasaCtrl}                & 28.38           & 22.17  & 106.62  & 86.97  & 79.67 & 23.96        & 21.16          \\
\midrule
\textbf{Ours}& \textbf{MasaCtrl}                & \textbf{24.70}\textcolor{red}{$_{13\%\downarrow}$}           & \textbf{22.64}\textcolor{red}{$_{2\%\uparrow}$}  & \textbf{87.94}\textcolor{red}{$_{18\%\downarrow}$}  & \textbf{81.09}\textcolor{red}{$_{7\%\downarrow}$}  & \textbf{81.33}\textcolor{red}{$_{2\%\uparrow}$} & \textbf{24.38}\textcolor{red}{$_{1.8\%\uparrow}$}        & \textbf{21.35}\textcolor{red}{$_{0.9\%\uparrow}$}          \\
\midrule
\midrule
\textbf{DDIM}& \textbf{P2P-Zero} & 61.68 & 20.44  & 172.22 & 144.12 & 74.67 & 22.80 & 20.54 \\
\midrule
\textbf{Ours}& \textbf{P2P-Zero} & \textbf{49.22}\textcolor{red}{$_{20\%\downarrow}$} & \textbf{21.53}\textcolor{red}{$_{5\%\uparrow}$} & \textbf{138.98}\textcolor{red}{$_{19\%\downarrow}$} & \textbf{127.32}\textcolor{red}{$_{12\%\downarrow}$} & \textbf{77.05}\textcolor{red}{$_{3\%\uparrow}$} & \textbf{23.31}\textcolor{red}{$_{2.2\%\uparrow}$} & \textbf{21.05}\textcolor{red}{$_{2.5\%\uparrow}$} \\
\midrule
\midrule
\textbf{DDIM}& \textbf{PnP\tnote{*}} & 28.22 & 22.28  & 113.46 & 83.64 & 79.05 & \textbf{25.41} & 22.55 \\
\midrule
\textbf{Ours}& \textbf{PnP\tnote{*}} & \textbf{24.29}\textcolor{red}{$_{14\%\downarrow}$} & \textbf{22.46}\textcolor{red}{$_{1\%\uparrow}$} & \textbf{106.06}\textcolor{red}{$_{7\%\downarrow}$} & \textbf{80.45}\textcolor{red}{$_{4\%\downarrow}$} & \textbf{79.68}\textcolor{red}{$_{1\%\uparrow}$} & \textbf{25.41} & \textbf{22.62}\textcolor{red}{$_{0.3\%\uparrow}$} \\
\bottomrule
\bottomrule
\end{tabular}
\begin{tablenotes}
   \footnotesize
   \item[*] use Stable Diffusion v1.5 as base model (others all use Stable Diffusion v1.4) %revise the source code's 1000 steps to 50 steps
   \item[\dag] averaged results on A800 and RTX3090 since different environment leads to slightly different performance
    \end{tablenotes}
    \end{threeparttable}}
    \vspace{-0.2cm}
\caption{\textbf{Compare \OursMethod with other inversion techniques across various editing methods.} For editing method Prompt-to-Prompt (P2P)~\citep{hertz2022prompt}, we compare four different inversion methods: DDIM Inversion (DDIM)~\citep{song2020denoising}, Null-Text Inversion (NT)~\citep{mokady2023null}, Negative-Prompt Inversion (NP)~\citep{miyake2023negative}, and StyleDiffusion (StyleD)~\citep{li2023stylediffusion}. For editing methods MasaCtrl~\citep{cao2023masactrl}, Pix2Pix-Zero (P2P-Zero)~\citep{cao2023masactrl}, Plug-and-Play (PnP)~\citep{tumanyan2023plug}, we compare with DDIM Inversion (DDIM).}
        \label{tab:inversion_based_editing}
    \vspace{-0.2cm}
\end{table}

\begin{figure}[htbp]
    \vspace{-0.6cm}
        \begin{center}
            \includegraphics[width=0.87\textwidth]{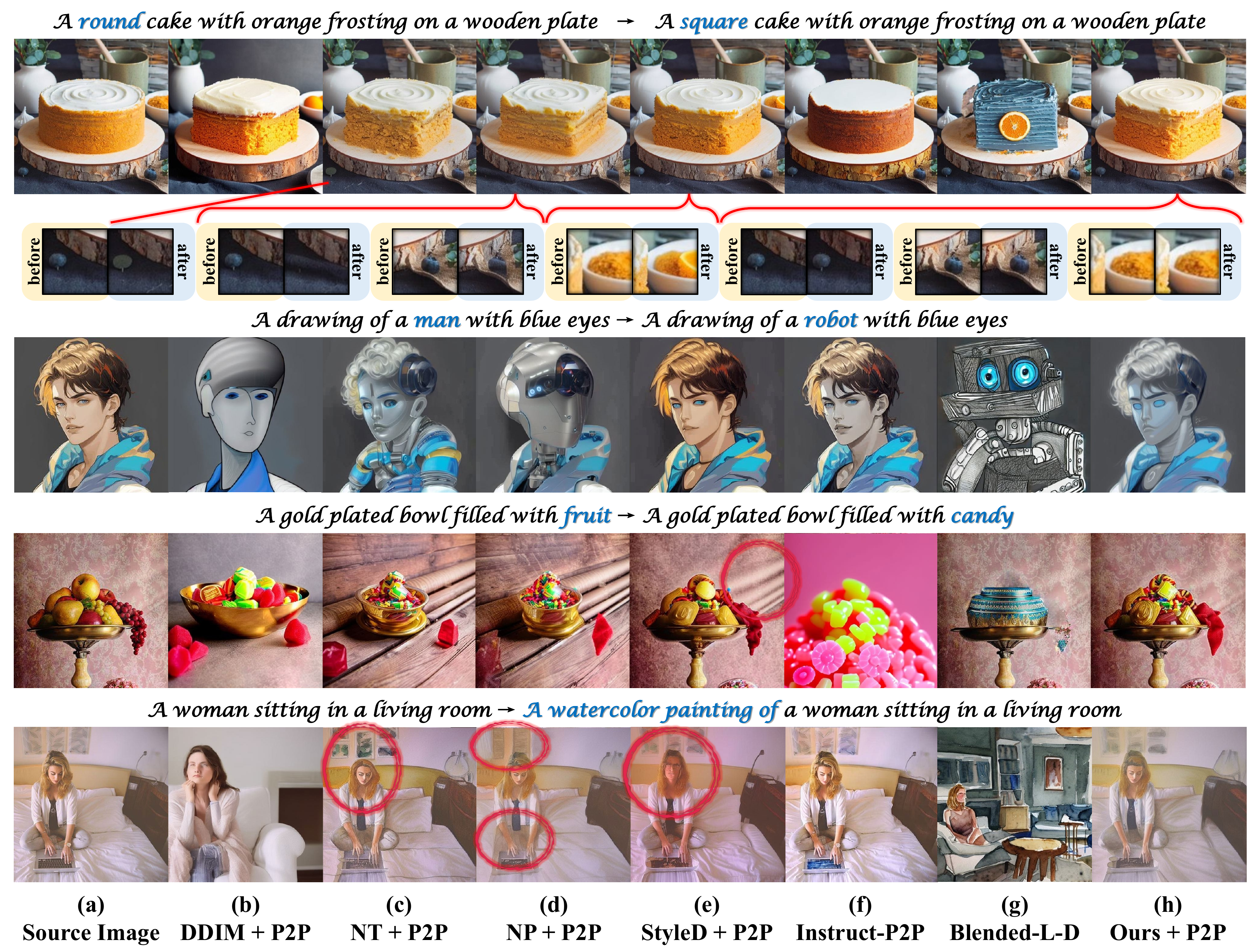}
        \end{center}
    \vspace{-0.45cm}
    \caption{\textbf{Visulization results of different inversion and editing techniques.} The source image is shown in col (a). We compare (h) \OursMethod with different inversion techniques added with Prompt-to-Prompt~\citep{hertz2022prompt}: (b) DDIM Inversion~\citep{song2020denoising}, (c) Null-Text Inversion~\citep{mokady2023null}, (d) Negative-Prompt Inversion~\citep{miyake2023negative}, and (e) StyleDiffusion~\citep{li2023stylediffusion}. We also compare model-based editing results: (f) Instruct-Pix2Pix~\citep{brooks2023instructpix2pix} and (g) Blended Latent Diffusion~\citep{avrahami2023blended}. The improvements are mostly tangible, and we circle some of the subtle discrepancies w/o \OursMethod in red.}
        \label{fig:p2p}
    \vspace{-0.65cm}
\end{figure}

\setlength{\tabcolsep}{2pt}
\begin{wraptable}{r}{0.23\textwidth}
 \vspace{-0.5cm}
	\centering
	\small
  \renewcommand\arraystretch{0.9}

	{%
		\begin{tabular}{c|c}
  
            \toprule
            Method & Time (s) \\
            \midrule
            NP & \textbf{18.22}  \\
            EF & 19.10   \\
            EDICT & 35.48   \\
            NT & 148.48  \\
            StyleD & 382.98   \\
            \midrule
            Ours & 28.17    \\
            \bottomrule
            \end{tabular}
	}
 \vspace{-0.25cm}
        \caption{\textbf{Inference time of different inversion techniques.}}
	\label{tab:inference_time}
 \vspace{-0.7cm}
\end{wraptable}

In this section, we compare \OursMethod with previous inversion-based editing methods quantitatively and qualitatively. Four inversion methods, DDIM Inversion~\citep{song2020denoising}, Null-Text Inversion~\citep{mokady2023null}, Negative-Prompt Inversion~\citep{miyake2023negative}, and StyleDiffusion~\citep{li2023stylediffusion}, as well as four editing methods, Prompt-to-Prompt~\citep{hertz2022prompt}, MasaCtrl~\citep{cao2023masactrl}, pix2pix-zero~\citep{parmar2023zero}, and Plug-and-Play~\citep{tumanyan2023plug} are taken into consideration. For inversion methods such as EDICT~\citep{wallace2023edict} and Edit-Friendly DDPM Inversion~\citep{huberman2023edit}, we put them into section~\ref{sec:Comparison_with_Background_Preservation_Methods} since their main target is to preserve the background and result in a decline in editability.

Table~\ref{tab:inversion_based_editing} shows the structure retention, background preservation, and edit clip similarity of the four inversion methods and the four editing methods. Results show that when added with \OursMethod, all editing methods have a performance improvement on the retention of background and structure while improving or maintaining editability compared with other inversion methods. While DDIM Inversion may yield a high edit CLIP Similarity within the edited mask, the preservation of structure and background falls significantly below the acceptable performance threshold, as depicted in Figure~\ref{fig:p2p}. We also give out the inference time of each inversion method added with Prompt-to-Prompt in Table~\ref{tab:inference_time}. \OursMethod reaches better editing results with far less inference time compared to Null-Text Inversion and StyleDiffusion. Although Negative-Prompt Inversion and Edit Friendly DDPM infer a little faster than \OursMethod, their editing results are much more unacceptable compared to \OursMethod as shown in Table~\ref{tab:inversion_based_editing} and Table~\ref{tab:background_preservation_methods}.

\vspace{-0.2cm}

\subsection{Comparison with Essential Content Preservation Methods}
\label{sec:Comparison_with_Background_Preservation_Methods}

\vspace{-0.2cm}

\setlength{\tabcolsep}{2pt}
\begin{wraptable}{r}{0.29\textwidth}
\vspace{-0.3cm}
\footnotesize
\centering
  \renewcommand\arraystretch{0.8}

\setlength{\tabcolsep}{0.4mm}{
\begin{tabular}{c|c|c|cccc|cc}
\toprule
\multicolumn{2}{c|}{\textbf{Method}}&  \multicolumn{2}{c}{\textbf{CLIP Similariy}} \\ \midrule
\textbf{Inverse} & \textbf{Edit}    & \textbf{Whole}  $\uparrow$    & \textbf{Edit}  $\uparrow$ \\ \midrule
\textbf{NT}& \textbf{P2P}     &  24.75        & 21.86      \\
\midrule
\textbf{NT} & \textbf{Prox}  &  22.91\textcolor{blue}{$\downarrow$} & 20.23\textcolor{blue}{$\downarrow$} \\
\textbf{NP} & \textbf{Prox}  &  24.28\textcolor{blue}{$\downarrow$} & 21.36\textcolor{blue}{$\downarrow$} \\
\textbf{EF} & \textbf{P2P}  &  23.97\textcolor{blue}{$\downarrow$} & 21.03\textcolor{blue}{$\downarrow$} \\
\textbf{EDICT} & \textbf{P2P}  &  23.09\textcolor{blue}{$\downarrow$} & 20.32\textcolor{blue}{$\downarrow$} \\
% \textbf{EDICT} & / &  24.46\textcolor{blue}{$\downarrow$} & 21.56\textcolor{blue}{$\downarrow$} \\
\midrule
\textbf{Ours}  & \textbf{P2P}     &  \textbf{25.02}\textcolor{red}{$\uparrow$} & \textbf{22.10}\textcolor{red}{$\uparrow$} \\
\bottomrule
\end{tabular}}

\vspace{-0.2cm}
\caption{\textbf{Compare \OursMethod with background preservation methods.}}
\label{tab:background_preservation_methods}
 \vspace{-0.2cm}
\end{wraptable}

We compare \OursMethod with inversion and editing techniques targeted for essential content preservation in Table~\ref{tab:background_preservation_methods}. Null-Text Inversion (NT)~\citep{mokady2023null} added with Prompt-to-Prompt (P2P)~\citep{hertz2022prompt} provides a baseline for all improvement methods. Specifically, Negative-Prompt Inversion (NP)~\citep{miyake2023negative} maintains a guidance scale of $1$ to reduce the deviation in editing. Proximal Guidance (Prox)~\citep{han2023improving} limits edit changes to a specific area based on editing amplitude. Edit Friendly DDPM (EF)~\citep{huberman2023edit} changes the DDPM sampling distribution to allow reconstruction of the desired image.
EDICT~\citep{wallace2023edict} maintains two coupled noise vectors to invert each other for image reconstruction. Although some of these techniques improve the structure and background preservation compared to Null-Text Inversion, clip similarity has decreased for all methods, which indicates a deteriorating editing ability. On the contrary, \OursMethod can lift structure/background preservation and editability simultaneously as shown in Table~\ref{tab:inversion_based_editing}.

\vspace{-0.2cm}

\subsection{Ablation Study}

\subsubsection{Comparing \OursMethod and Null-Text Inversion}

To validate our theoretical analysis, we prove experimentally in Table~\ref{tab:ablation_null_text} that \OursMethod's improvement over Null-Text Inversion (NT) is a three-step process, disentangling the source and target branch, wiping off the force assignment of null-text embedding, and removing the distance gap shown in Figure~\ref{fig:inverse}. To disentangle the two branches, we revise Null-Text Inversion to a single-branch version (NT-S) and only assign the learned null-text latent to the source branch. Results show an improvement in CLIP Similarity, revealing the benefit of leaving the target branch unaltered. To wipe off the force assignment, we use the optimization strategy of Null-Text Inversion, and instead of replacing null-text embedding, we directly add the difference to the source latent. The result is shown as Null-Latent Inversion (NL). To show the influence of the distance gap, we add scaled distance ($scale*o_{t}^{src}$) to the source latent. Results show that with the distance gap increase, the structure and background preservation decline, while the edit fidelity fluctuates. Moreover, the Null-Latent Inversion's performance is between added distance with a scale of 0.4 and 0.8, which implies the average optimization distance gap of Null-Text inversion is between 0.4 and 0.8.

\begin{table}[htbp]
\footnotesize
\centering
\setlength{\tabcolsep}{1.6mm}{
  \renewcommand\arraystretch{0.8}
\begin{threeparttable}
\begin{tabular}{c|c|cccc|cc}
\toprule
\textbf{Metrics}        & \textbf{Structure}          & \multicolumn{4}{c|}{\textbf{Background Preservation}} & \multicolumn{2}{c}{\textbf{CLIP Similarity}} \\ \midrule
\textbf{Method}            & \textbf{Distance}$_{^{\times 10^3}}$ $\downarrow$ & \textbf{PSNR} $\uparrow$     & \textbf{LPIPS}$_{^{\times 10^3}}$ $\downarrow$  & \textbf{MSE}$_{^{\times 10^4}}$ $\downarrow$     & \textbf{SSIM}$_{^{\times 10^2}}$ $\uparrow$    & \textbf{Whole}  $\uparrow$          & \textbf{Edited}  $\uparrow$       \\ \midrule
\textbf{NT\tnote{\dag}}     & 13.44           & 27.03  & 60.67  & 35.86  & 84.11 & 24.75        & 21.86      \\
\textbf{NT-S\tnote{\dag}}     & 14.25\textcolor{blue}{$\uparrow$} & 26.39\textcolor{blue}{$\downarrow$} & 66.62\textcolor{blue}{$\uparrow$} & 40.09\textcolor{blue}{$\uparrow$} & 83.52\textcolor{blue}{$\downarrow$} & 25.01\textcolor{red}{$\uparrow$}  & 22.11\textcolor{red}{$\uparrow$}      \\
\midrule
\textbf{Scale}            & \textbf{Distance}$_{^{\times 10^3}}$ $\downarrow$ & \textbf{PSNR} $\uparrow$     & \textbf{LPIPS}$_{^{\times 10^3}}$ $\downarrow$  & \textbf{MSE}$_{^{\times 10^4}}$ $\downarrow$     & \textbf{SSIM}$_{^{\times 10^2}}$ $\uparrow$    & \textbf{Whole}  $\uparrow$          & \textbf{Edited}  $\uparrow$       \\ \midrule
\textbf{0.4} & 13.55 & 26.65 & 58.79 & 36.98 & 84.29 & 25.02 & 22.10 \\ 
\textbf{NL\tnote{\dag}} & 12.05\textcolor{red}{$\downarrow$} & 27.03\textcolor{red}{$\uparrow$} & 55.83\textcolor{red}{$\downarrow$} & 33.94\textcolor{red}{$\downarrow$} & 84.55\textcolor{red}{$\uparrow$} & 25.02\textcolor{red}{$\uparrow$} & 22.09\textcolor{red}{$\uparrow$}     \\ 
\textbf{0.8} & 11.90 & 27.14 & 54.76 & 33.35 & 84.66 & \textbf{25.08} & \textbf{22.12} \\ 
\midrule
\textbf{1}                & \textbf{11.65}\textcolor{red}{$\downarrow$} & \textbf{27.22}\textcolor{red}{$\uparrow$} & \textbf{54.55}\textcolor{red}{$\uparrow$} & \textbf{32.86}\textcolor{red}{$\downarrow$} & \textbf{84.76}\textcolor{red}{$\uparrow$} & 25.02\textcolor{red}{$\uparrow$} & 22.10\textcolor{red}{$\uparrow$}        \\

\bottomrule
\end{tabular}
\begin{tablenotes}
   \footnotesize
   \item[\dag] averaged results on A800 and RTX3090 since different environment leads to slightly different performance
    \end{tablenotes}
    \end{threeparttable}}

\caption{\textbf{Ablation study of comparing Null-Text Inversion and \OursMethod.}}
\label{tab:ablation_null_text}
\vspace{-0.2cm}
\end{table}

\vspace{-0.1cm}

\subsubsection{Influence of Adding Difference to Target Latent}

In Algorithm~\ref{alg:direct_inversion}, we only add the distance of the source prompt to the source latent. To show the rationality of this operation and the disentanglement of the source and target branch, we compare the performance of adding source distance to the target latent and adding target distance to the target latent in Table~\ref{tab:add_target}. Adding source distance to the target latent leads to a decline in both structure/background preservation and clip similarity. Although adding target distance to the target latent leads to better structure/background preservation, the clip similarity (edit fidelity) sharply decreases.

\begin{table}[htbp]
\footnotesize
\centering
  \renewcommand\arraystretch{0.8}
\setlength{\tabcolsep}{0.4mm}{
\begin{tabular}{c|c|cccc|cc}
\toprule
\textbf{Metrics}     & \textbf{Structure}          & \multicolumn{4}{c|}{\textbf{Background Preservation}} & \multicolumn{2}{c}{\textbf{CLIP Similariy}} \\ \midrule 
\textbf{Add} & \textbf{Distance}$_{^{\times 10^3}}$ $\downarrow$ & \textbf{PSNR} $\uparrow$     & \textbf{LPIPS}$_{^{\times 10^3}}$ $\downarrow$  & \textbf{MSE}$_{^{\times 10^4}}$ $\downarrow$     & \textbf{SSIM}$_{^{\times 10^2}}$ $\uparrow$    & \textbf{Whole}  $\uparrow$          & \textbf{Edit}  $\uparrow$       \\ \midrule
$o_{t-1}^{src}$  & 19.30 & 26.15 & 63.70 & 46.45 & 83.67 & 24.93 & 21.88  \\
$o_{t-1}^{tgt}$ & \textbf{9.86} & \textbf{27.66} & \textbf{50.17} & \textbf{33.28} & \textbf{85.13} & 23.00 & 20.27 \\
0             & 11.65 & 27.22 & 54.55 & 32.86 & 84.76 & \textbf{25.02} & \textbf{22.10}   \\
\bottomrule
\end{tabular}}
\vspace{-0.2cm}
\caption{\textbf{Results of adding the difference to the target latent.}}
\label{tab:add_target}
\vspace{-0.2cm}
\end{table}

\vspace{-0.2cm}

\vspace{-0.2cm}
\section{Conclusion}\label{sec:conclusion}
\vspace{-0.2cm}

This paper introduces \OursMethod, a simple yet effective technique for inverting diffusion models. By disentangling the source and target branches in diffusion-based editing, \OursMethod separates the responsibility for essential content preservation and edit fidelity, thus achieving superior performance in both aspects. To address the lack of standardized performance criteria for inversion and editing techniques, we develop \Benchmark comprising 700 images in natural and artificial scenes featuring ten distinct editing types. Evaluation metrics demonstrate that \OursMethod outperforms eight editing methods across five inversion techniques in terms of both edit quality and inference speed. Limitations and future work can be found in supplementary files.

\bibliography{iclr2024_conference}
\bibliographystyle{iclr2024_conference}

\appendix

\newpage

\textbf{Reproducibility Statement.} To ensure the reproducibility and completeness of this paper, we include the Appendix with $7$ sections. Appendix~\ref{supp:related_work} provides details of related works, offering additional information to complement the main text. Appendix~\ref{supp:benchmark_construction} introduces the construction of \Benchmark in detail and provides examples in the benchmark. Appendix~\ref{supp:evaluation_metrics} illustrates the evaluation metrics we use in our experiments. Appendix~\ref{supp:implementation_details} contains the details of our implementation. Appendix~\ref{supp:quantitative_results} contains quantitative results on the reconstruction ability of different inversion methods, a full comparison with essential content preservation methods, a comparison with model-based editing, ablation of step and interval, influence of inverse and forward guidance scale, and results of different editing types. Appendix~\ref{supp:qualitative_results} provides more qualitative results compared with different inversion-based editing, essential content preservation methods, and model-based editing. Lastly, we include limitations and future works in Section~\ref{supp:limitation}.

\section{Related Work}
\label{supp:related_work}

Diffusion models have shown exceptional realism and diversity~\citep{yu2023freedom,ju2023humansd,chen2023humanmac,wang2022zero} in computer vision tasks, including image generation and editing. As mentioned in the main paper, diffusion-based image editing involves two primary concerns: (1) edit fidelity and (2) essential content preservation. 
Most diffusion-based editing methods take both aspects into consideration and perform editing using a two branches strategy, \emph{i.e.}, a source diffusion branch to maintain the source image's essential content and a target diffusion branch to insert editing instruction, as shown in Figure~\ref{fig:related_work}.
Accordingly, we undertake a comprehensive review of prior methodologies concerning both two aspects.

\paragraph{Edit Fidelity.}

Diffusion models naturally possess hierarchical features (\emph{e.g.}, noisy latent of each step,  different resolution UNet features), enabling different editing strategies.
Previous methods perform editing roughly through four ways: 
\ding{172} end-to-end editing model (with only one editing branch)
, \ding{173} latent integration
, \ding{174} attention integration
, and \ding{175} target embedding
.

Specifically, \ding{172} trains end-to-end diffusion models for image editing, which is limited by insufficient/noisy training data or indirect training strategies~\citep{brooks2023instructpix2pix, kim2022diffusionclip, nichol2021glide,mirzaei2023watch}.
The shared objective of \ding{173}-\ding{175} is to map both the source image and the target editing instruction to the diffusion space, then inject the target branch's features into the source diffusion space. \ding{173} inserts editing instruction in the level of the noisy diffusion latent~\citep{meng2022sdedit, avrahami2022blended, avrahami2023blended, couairon2022diffedit, zhang2023sine, joseph2023iterative}.
The features of the source and the target branch are merged through mask stitching~\citep{meng2022sdedit, avrahami2022blended, avrahami2023blended, couairon2022diffedit, joseph2023iterative} or weighted addition~\citep{zhang2023sine, shi2023dragdiffusion}.
However, using mask stitching for feature insertion may lead to abrupt editing boundaries, and using weighted addition makes it difficult to make refined modifications. 
\ding{174} tries to solve these two problems by fusing in a more refined feature space, the attention map that connects the text and image. 
Prompt-to-prompt~\citep{hertz2022prompt} directly replaces the cross-attention map to perform editing through text.
Proximal guidance~\citep{han2023improving}, Zero-shot~\citep{parmar2023zero}, MasaCtrl~\citep{cao2023masactrl}, Plug-and-Play~\citep{tumanyan2023plug}, RIVAL~\citep{zhang2023real}, and DragonDiffusion~\citep{mou2023dragondiffusion} further extend the use of both cross-attention and self-attention map to achieve better editing results or explore more applications.
\ding{175} first aggregates editing information into an embedding, then uses this embedding to perform editing on the source diffusion branch, which may confront long feature extraction times and unstable editing performance~\citep{kawar2023imagic, cheng2023general, wu2023uncovering, brack2023sega, tsaban2023ledits, valevski2022unitune, dong2023prompt}.

\paragraph{Essential Content Preservation.} 

While the methods mentioned enable basic image editing, preserving the essential content, particularly on images devoid of inherent diffusion space, remains challenging. Previous methods tried to solve this problem through \ding{182} overfit the editing image, \ding{183} DDPM/DDIM inversion variation, \ding{184} attention preservation, and \ding{185} source embedding.

Specifically, \ding{182} overfits the source image to avoid significant image variation~\citep{kawar2023imagic,shi2023dragdiffusion}. \ding{183} makes variations on the DDPM/DDIM sampling process to adapt the editing. Negative-prompt Inversion~\citep{miyake2023negative} set the classifier-free guidance scale to 1 to reduce the deviation caused by DDIM inversion, which weakens the text's controllability. Edit Friendly Noise~\citep{huberman2023edit} imprints the source image more strongly onto the noise space to ensure better reconstruction. However, this reduces the modification space due to the decrease in noise. EDICT~\citep{wallace2023edict} maintains two coupled noise vectors to reach mathematically exact inversion, but leading to a decrease of edit fidelity. \ding{184} devises ways of utilizing both cross-attention and self-attention map with better balance of semantic editing results and original image structure~\citep{mou2023dragondiffusion,cheng2023general,cao2023masactrl,parmar2023zero,tumanyan2023plug,hertz2022prompt,qi2023fatezero}. \ding{185} absorbs source image to an embedding and use this embedding to reconstruct the essential content of the source image~\citep{mokady2023null, dong2023prompt, li2023stylediffusion,gal2022image,fei2023gradient,huang2023reversion}. Specifically, Null-text inversion~\citep{huang2023reversion} optimizes a Null embedding to capture the difference between the reconstructed image and the source image. Subsequently, this difference is steply reintroduced in both source and target branch during the editing procedure. However, null-text inversion necessitates prolonged optimization times per image, lacks the assurance of achieving flawless optimization, and disturbs the diffusion model distribution. Prompt Tuning Inversion~\citep{dong2023prompt} and StyleDiffusion~\citep{li2023stylediffusion} optimize text embedding and cross-attention value to capture the difference instead of null-text, thus facing the same issue with Null-text Inversion.

\begin{table}[htbp]
\small
\centering
\setlength{\tabcolsep}{0.45mm}{
\begin{tabular}{c|cccc|cccc|c|cccc|cccc}
\toprule
\multirow{2}{*}{\textbf{Method}} & \multicolumn{4}{c|}{\begin{tabular}[c]{@{}c@{}}\textbf{Source}\\ \textbf{Branch}\end{tabular}} & \multicolumn{4}{c|}{\begin{tabular}[c]{@{}c@{}}\textbf{Target}\\ \textbf{Branch}\end{tabular}} & \multirow{2}{*}{\textbf{Method}} & \multicolumn{4}{c|}{\begin{tabular}[c]{@{}c@{}}\textbf{Source}\\ \textbf{Branch}\end{tabular}} & \multicolumn{4}{c}{\begin{tabular}[c]{@{}c@{}}\textbf{Target}\\ \textbf{Branch}\end{tabular}} \\ \cmidrule{2-9} \cmidrule{11-18} & \ding{182}   & \ding{183}  & \ding{184}  & \ding{185}  & \ding{172} & \ding{173} & \ding{174} & \ding{175} &   & \ding{182}   & \ding{183}  & \ding{184}  & \ding{185}  & \ding{172} & \ding{173} & \ding{174} & \ding{175}     \\ 
\midrule
    \cite{mokady2023null} &  &    &    &  \checkmark  &   &   &   &   &
    \cite{cao2023masactrl} & &    & \checkmark &    &  &  & \checkmark &  
    \\
    \cite{mou2023dragondiffusion}  &  &    &  \checkmark  &    &   &  \checkmark &  \checkmark &   &
    \cite{brooks2023instructpix2pix} & \checkmark &  &  &  & \checkmark &  &   &  
    \\
    \cite{kim2022diffusionclip} & \checkmark & \checkmark &  &  & \checkmark &  &   &   &
    \cite{nichol2021glide} & \checkmark &  &  &  & \checkmark &  &  &  
    \\ 
    \cite{mirzaei2023watch} &  &  &  &  & \checkmark &  &  &  &
    \cite{meng2022sdedit} &  & &  &  &  & \checkmark &  &  
    \\ 
    \cite{avrahami2023blended} &  &  &  &  &  & \checkmark &  &  &
    \cite{avrahami2022blended} &  &  &  &  &  & \checkmark &  &  
    \\ 
    \cite{couairon2022diffedit} &  &  &  &  &  & \checkmark &  &  &
    \cite{zhang2023sine} & \checkmark &  &  &  &  & \checkmark &  &  
    \\ 
    \cite{shi2023dragdiffusion} & \checkmark &  &  &  &  & \checkmark &  &  &
    \cite{hertz2022prompt} &  &  & \checkmark &  &  &  & \checkmark &  
    \\ 
    \cite{han2023improving} &  & \checkmark &  &  &  &  & \checkmark &  &
    \cite{parmar2023zero} &  &  & \checkmark &  &  &  & \checkmark &  
    \\ 
    \cite{tumanyan2023plug} &  &  & \checkmark &  &  &  & \checkmark &  &
    \cite{zhang2023real} &  & \checkmark & \checkmark &  &  &  & \checkmark &  
    \\ 
    \cite{kawar2023imagic} & \checkmark &  &  &  &  &  &  & \checkmark &
    \cite{cheng2023general} &  &  & \checkmark  &  &  &  &  & \checkmark
    \\ 
    \cite{wu2023uncovering} &  &  &  & \checkmark &  &  &  & \checkmark &
    \cite{brack2023sega} &  &  &  &  &  &  &  & \checkmark 
    \\ 
    \cite{tsaban2023ledits} &  &  &  &  &  &  &  & \checkmark &
    \cite{valevski2022unitune} & \checkmark &  &  &  &  &  &  & \checkmark 
    \\ 
    \cite{dong2023prompt} &  &  &  & \checkmark &  &  &  & \checkmark &
    \cite{miyake2023negative} &  & \checkmark  &  &  &  &  &  & 
    \\ 
    \cite{huberman2023edit} &  & \checkmark  &  &  &  &  &  &  &
    \cite{qi2023fatezero} &  &  & \checkmark &  &  &  & \checkmark & 
    \\ 
    \cite{li2023stylediffusion} &  &  &  & \checkmark &  &  & \checkmark &  &
    \cite{gal2022image} &  &  &  & \checkmark &  &  &  & \checkmark
    \\  
    \cite{fei2023gradient} &  &  &  & \checkmark &  &  &  &  &
    \cite{huang2023reversion} &  &  &  & \checkmark &  &  &  & 
    \\  
    \cite{wallace2023edict} &  & \checkmark &  &  &  &  &  &  &
    \cite{joseph2023iterative} &  &  &  &  &  & \checkmark &  & 
    \\
    \cite{geng2023instructdiffusion} &  &  &  & \checkmark &  &   & 
     &  &  &  &  &  &  &  & 
    \\
    \bottomrule
\end{tabular}}
\caption{\textbf{Strategies for enhancing editing fidelity and preserving essential content in previous diffusion-based editing methods.}}
\label{tab:related_work}
\end{table}

More refined categorization is presented in Table~\ref{tab:related_work}. To summarize, existing background preservation methods suffer from unstable and time-consuming optimization processes, as well as the persisting issue of error propagation inversion. Moreover, the absence of a disentanglement for the source and target branches is unfavorable for achieving optimal performance in both edit fidelity and essential content preservation. Instead, a simple yet effective \OursMethod is capable of achieving superior results with virtually negligible computational cost and negligible inversion error without optimization by branch disentanglement, aiding in accurately editing the real images while preserving the structural information.

\section{Benchmark Construction}
\label{supp:benchmark_construction}

Although diffusion-based editing has been widely explored in recent years, people mainly evaluate the performance of different editing methods with subjective and incomprehensive visualization results. Previously, PnP~\citep{tumanyan2023plug} provides a benchmark of 55 images with editing prompts. Instruct-Pix2Pix~\citep{brooks2023instructpix2pix} builds a dataset with randomly selected 451,990 images, editing prompts written by ChatGPT, and pseudo editing results of Null-Text Inversion~\citep{mokady2023null} and Prompt2Prompt~\citep{hertz2022prompt}. However, without manual labels and fine-grained classification, these datasets are not capable of supporting comprehensive metrics evaluation.

To systematically validate our proposed method as a plug-and-play strategy for editing models and compare our method with existing inversion methods, as well as compensate for the absence of standardized performance criteria for inversion and editing techniques, we construct a benchmark dataset, named \Benchmark (\textbf{P}rompt-based \textbf{I}mage \textbf{E}diting \textbf{Bench}mark).

\begin{figure}[htbp]
        \begin{center}
            \includegraphics[width=1\textwidth]{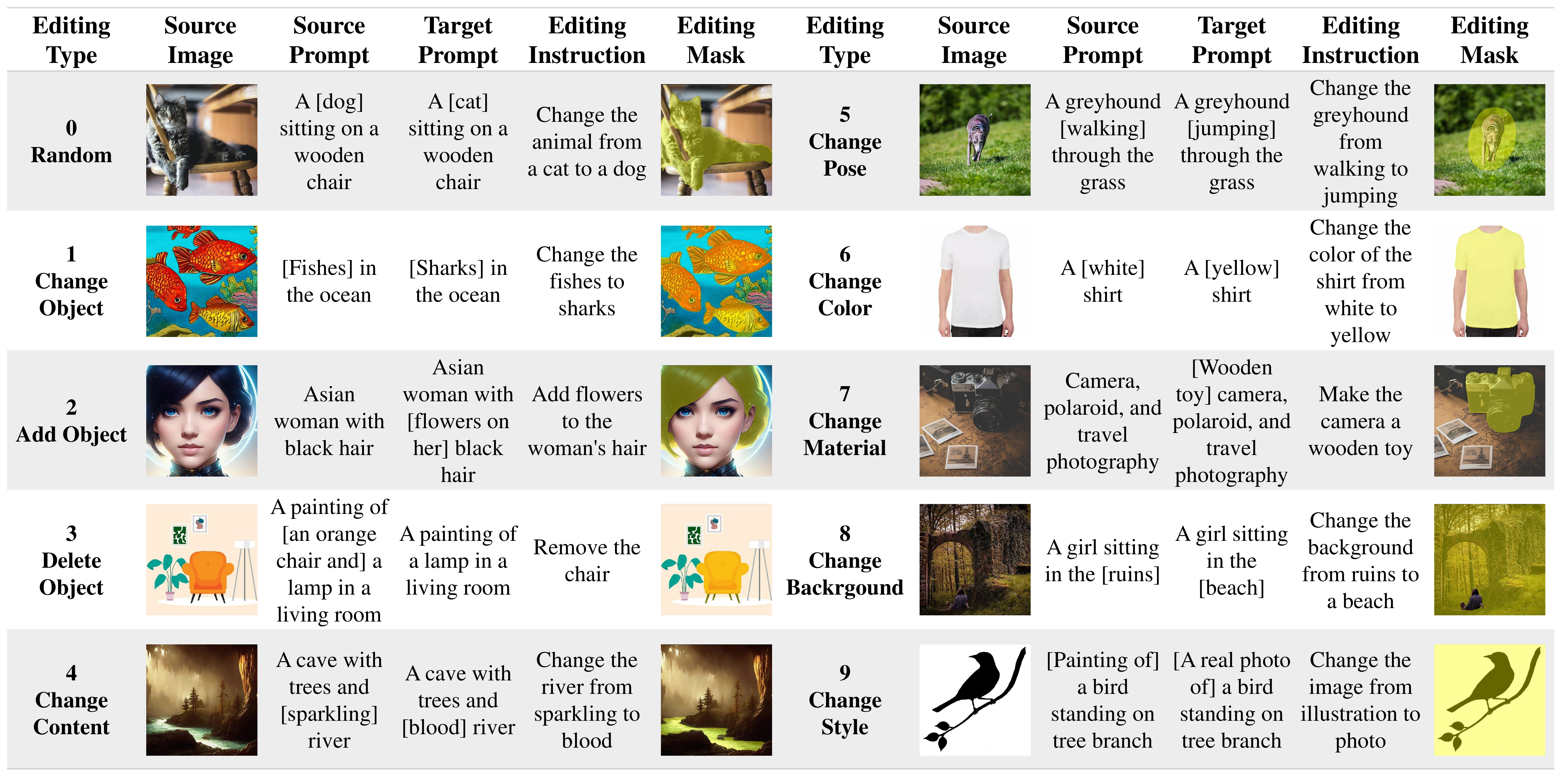}
        \end{center}
    \caption{\textbf{Examples of \Benchmark}. One example is provided for each editing type.}
        \label{fig:benchmark}
\end{figure}

\Benchmark comprises 700 images in natural and artificial scenes (\emph{e.g.}, paintings) featuring ten distinct editing types as shown in Figure~\ref{fig:benchmark}: (0) random editing written by volunteers, (1) change object, (2) add object, (3) delete object, (4) change object content, (5) change object pose, (6) change object color, (7) change object material, (8) change background, and (9) change image style. In each editing type of 1-9, images are evenly distributed among natural and artificial scenes. Within each scene, images are evenly distributed among four categories: animal, human, indoor environment, and outdoor environment.
Each image in \Benchmark includes five annotations: a source image prompt, a target image prompt, an editing instruction, edit subjects describing the main editing body, and the editing mask. For editing type 0, we invited some volunteers to write the source image prompt, target image prompt, and editing instructions based on their editing expectations. For the other editing types, we employ BLIP-2~\citep{li2023blip} to generate the source image prompt and use GPT4~\citep{OpenAI2023GPT4TR} to craft the target image prompt and editing instructions tailored to each editing type. Then, we manually rectify any inaccuracies in the automatically generated captions, target prompt, and edit instructions. Subsequently, 2 data annotators and 1 data auditor annotate the main editing body as well as the editing mask (indicating the anticipated editing region) in an image. Notably, the editing mask annotation is crucial in accurate metrics computations as we expect the editing to only occur within the designated area.

\section{Evaluaion Metrics}
\label{supp:evaluation_metrics}

To illustrate the effectiveness and efficiency of our proposed \OursMethod, we use eight metrics covering four aspects: structure distance, background preservation, edit prompt-image consistency and inference time.

\textbf{Structure Distance:} We follow ~\cite{tumanyan2022splicing} to leverage self-similarity of deep spatial features extracted from DINO-ViT as a structure representation and use cosine similarity between image features as structure distance. The structure distance can capture structure while ignoring appearance information. Thus, it is well-suited for our proposed benchmark and diffusion-based editing methods since we do not expect a huge structural change.

\textbf{Background Preservation:} We calculate standard PSNR, LPIPS~\citep{zhang2018unreasonable}, MSE, and SSIM~\citep{wang2004image} in the area outside of the manual-annotated masks of \Benchmark to demonstrate the background preservation ability of different inversion and editing techniques. 

\textbf{Edit Text-image Consistency:} The CLIP~\citep{clip} Similarity (CLIPSIM~\citep{clipsim}) evaluates text-image consistency between the edited images and corresponding target editing text prompts. CLIP Similarity projects text and images to the same shared space and evaluates the similarity of their embeddings. We calculate CLIP Similarity both on the whole image and in the editing mask (black out everything outside the mask) to demonstrate the performance of editing, as well as reflecting the editability. These two metrics are called Whole Image Clip Similarity and Edit Region Clip Similarity, respectively. 

\textbf{Inference Time:} We test inference time per image of different inversion techniques and Prompt-to-Prompt~\citep{hertz2022prompt} on one NVIDIA A800 80G to evaluate efficiency. Results are averaged over $20$ random runs.

\section{Implementation Details}
\label{supp:implementation_details}

We perform the inference of different editing and inversion methods in the same setting unless specifically clarified, \emph{i.e.}, on RTX3090 following their open-source code with a base model of Stabe Diffusion v1.4 in 50 steps, with an inverse guidance scale of 1 and a forward guidance scale of 0. Different images may have different hyper-parameters in different editing models, and we keep the recommended hyper-parameter for each editing method in all images for fair comparison. Details can be found in the provided code.

\section{Quantitative Results}
\label{supp:quantitative_results}

\subsection{Reconstruction Ability of Different Inversion Methods}

To further show the reconstruction ability of different inversion methods, we evaluate the reconstruction results of DDIM Inversion, Null-Text Inversion, Negative-Prompt Inversion, StyleDiffusion, and \OursMethod by giving source prompt as model input. We provide results of Structure Distance and Background preservation to show the ability to correct $z_{t}^{''}$ back to $z_{t}^{*}$. As shown in Table~\ref{tab:reconstruction}, \OursMethod is better than all these inversion methods on all metrics.

\begin{table}[htbp]
\footnotesize
\centering
\setlength{\tabcolsep}{2.9mm}{
\begin{threeparttable}
\begin{tabular}{c|c|cccc}
\toprule
\textbf{Inverse}           & \textbf{Structure Distance}$_{^{\times 10^3}}$ $\downarrow$ & \textbf{PSNR} $\uparrow$     & \textbf{LPIPS}$_{^{\times 10^3}}$ $\downarrow$  & \textbf{MSE}$_{^{\times 10^4}}$ $\downarrow$     & \textbf{SSIM}$_{^{\times 10^2}}$ $\uparrow$    \\ \midrule
\textbf{DDIM} & 70.23 & 17.76 & 210.84 & 224.43 & 70.96   \\
\textbf{NT\dag} & \underline{3.30} & \underline{30.17} & \underline{33.39} & \underline{18.86} & \underline{86.84}  \\
\textbf{NP} & 8.47 & 27.73 & 57.04 & 30.05 & 84.59   \\
\textbf{StyleD}&  4.35 & 28.88 & 39.45 & 22.63 & 86.07    \\
\midrule
\textbf{Ours} & \textbf{2.95} & \textbf{30.57} & \textbf{31.41} & \textbf{17.60} & \textbf{87.20}   \\
\bottomrule
\end{tabular}\begin{tablenotes}
   \footnotesize
   \item[\dag] averaged results on A800 and TRX3090 since different environment leads to slightly different performance
    \end{tablenotes}
    \end{threeparttable}}
\caption{\textbf{Reconstruction results of different inversion techniques}}
        \label{tab:reconstruction}
\end{table}

\subsection{Comparison with Essential Content Preservation Methods}

\begin{table}[htbp]
\footnotesize
\centering
\setlength{\tabcolsep}{0.4mm}{
\begin{tabular}{c|c|c|cccc|cc}
\toprule
\multicolumn{2}{c|}{\textbf{Method}}& \textbf{Structure}    & \multicolumn{4}{c|}{\textbf{Background Preservation}} & \multicolumn{2}{c}{\textbf{CLIP Similariy}} \\ \midrule
\textbf{Inverse}    & \textbf{Editing} & \textbf{Distance}$_{^{\times 10^3}}$ $\downarrow$ & \textbf{PSNR} $\uparrow$     & \textbf{LPIPS}$_{^{\times 10^3}}$ $\downarrow$  & \textbf{MSE}$_{^{\times 10^4}}$ $\downarrow$     & \textbf{SSIM}$_{^{\times 10^2}}$ $\uparrow$    & \textbf{Whole}  $\uparrow$    & \textbf{Edit}  $\uparrow$ \\ \midrule
\textbf{NT}& \textbf{P2P}     & 13.44           & 27.03  & 60.67  & 35.86  & 84.11 & 24.75        & 21.86      \\
\midrule
\textbf{NT} & \textbf{Prox}  & \textbf{3.51}\textcolor{red}{$\downarrow$} & \textbf{30.21}\textcolor{red}{$\uparrow$} & \textbf{32.97}\textcolor{red}{$\downarrow$} & \textbf{18.47}\textcolor{red}{$\downarrow$} & \textbf{87.01}\textcolor{red}{$\uparrow$} & 22.91\textcolor{blue}{$\downarrow$} & 20.23\textcolor{blue}{$\downarrow$} \\
\textbf{NP} & \textbf{Prox}  & 7.44\textcolor{red}{$\downarrow$} & 28.67\textcolor{red}{$\uparrow$} & 41.98\textcolor{red}{$\downarrow$} & 24.25\textcolor{red}{$\downarrow$} & 85.98\textcolor{red}{$\uparrow$} & 24.28\textcolor{blue}{$\downarrow$} & 21.36\textcolor{blue}{$\downarrow$} \\
\textbf{EF} & \textbf{P2P}  & 18.05\textcolor{blue}{$\uparrow$} & 24.55\textcolor{blue}{$\downarrow$} & 91.88\textcolor{blue}{$\uparrow$} & 94.58\textcolor{blue}{$\uparrow$} & 81.57\textcolor{blue}{$\downarrow$} & 23.97\textcolor{blue}{$\downarrow$} & 21.03\textcolor{blue}{$\downarrow$} \\
\textbf{EDICT} & \textbf{P2P}  & 4.61\textcolor{red}{$\downarrow$} & 29.79\textcolor{red}{$\uparrow$} & 37.03\textcolor{red}{$\downarrow$} & 20.37\textcolor{red}{$\downarrow$} & 86.55\textcolor{red}{$\uparrow$} & 23.09\textcolor{blue}{$\downarrow$} & 20.32\textcolor{blue}{$\downarrow$} \\
\textbf{EDICT} & / & 13.28\textcolor{red}{$\downarrow$} & 26.76\textcolor{blue}{$\downarrow$} & 65.51\textcolor{blue}{$\uparrow$} & 38.14\textcolor{blue}{$\uparrow$} & 83.72\textcolor{blue}{$\downarrow$} & 24.46\textcolor{blue}{$\downarrow$} & 21.56\textcolor{blue}{$\downarrow$}
 \\
\midrule
\textbf{Ours}  & \textbf{P2P}     & 11.65\textcolor{red}{$\uparrow$} & 27.22\textcolor{red}{$\uparrow$} & 54.55\textcolor{red}{$\uparrow$} & 32.86\textcolor{red}{$\uparrow$} & 84.76\textcolor{red}{$\uparrow$} & \textbf{25.02}\textcolor{red}{$\uparrow$} & \textbf{22.10}\textcolor{red}{$\uparrow$} \\
\bottomrule
\end{tabular}}
\vspace{-0.2cm}
\caption{\textbf{Full table of comparing \OursMethod with background preservation methods.} Null-Text Inversion (NT)~\citep{mokady2023null} added with Prompt-to-Prompt (P2P)~\citep{hertz2022prompt} provides a baseline for all improvement methods. Specifically, Negative-Prompt Inversion (NP)~\citep{miyake2023negative} maintains a guidance scale of $1$ to reduce the deviation in editing. Proximal Guidance (Prox)~\citep{han2023improving} limits edit changes to a specific area based on editing amplitude. Edit Friendly DDPM (EF)~\citep{huberman2023edit} changes the DDPM sampling distribution to allow reconstruction of the desired image.
EDICT~\citep{wallace2023edict} maintains two coupled noise vectors to invert each other for image reconstruction. }
\label{tab:background_preservation_methods_full}
\end{table}

We provide the full table of comparing \OursMethod with inversion and editing techniques targeted for background preservation in Table~\ref{tab:background_preservation_methods_full}. As explained in the main paper, although some of these techniques improve the structure and background preservation compared to Null-Text Inversion, clip similarity has decreased for all methods, which indicates a deteriorating editing ability. However, \OursMethod can lift structure/background preservation and editability simultaneously, which shows the effectiveness of \OursMethod in image editing.

\vspace{-0.3cm}

\subsection{Comparison with Model-based Editing}

We also compare three model-based editing methods, InstructPix2Pix~\citep{brooks2023instructpix2pix}, InstructDiffusion~\citep{geng2023instructdiffusion}, and Blended Latent Diffusion~\citep{avrahami2023blended} in Table~\ref{tab:end_to_end}. \OursMethod added with Prompt-to-Prompt shows a better structure and background preservation as well as better CLIP similarity than the two end-to-end editing models InstructPix2Pix and InstructDiffusion. Blended Diffusion uses an explicit mask and only performs editing in the mask. We directly use ground-truth mask in \Benchmark as input, which leads to better background preservation and CLIP similarity score. However, the forced editing makes the editing part incompatible with the background, as shown in Figure~\ref{fig:p2p}, and thus having a much larger Structure Distance compared to \OursMethod added with Prompt-to-Prompt.

\begin{table}[htbp]
\footnotesize
\centering
\setlength{\tabcolsep}{0.4mm}{
\begin{tabular}{c|c|cccc|cc}
\toprule
\textbf{Metrics}        & \textbf{Structure}          & \multicolumn{4}{c|}{\textbf{Background Preservation}} & \multicolumn{2}{c}{\textbf{CLIP Similariy}} \\ \midrule
\textbf{Method}            & \textbf{Distance}$_{^{\times 10^3}}$ $\downarrow$ & \textbf{PSNR} $\uparrow$     & \textbf{LPIPS}$_{^{\times 10^3}}$ $\downarrow$  & \textbf{MSE}$_{^{\times 10^4}}$ $\downarrow$     & \textbf{SSIM}$_{^{\times 10^2}}$ $\uparrow$    & \textbf{Whole}  $\uparrow$          & \textbf{Edit}  $\uparrow$       \\ \midrule
\textbf{InstructPix2Pix} & \underline{57.91}            & 20.82  & 158.63  & 227.78  & 76.26 & 23.61         & 21.64          \\ 
\textbf{InstructDiffusion} & 75.44         & 20.28  & 155.66  & 349.66  & 75.53 & 23.26         & 21.34          \\ 
\textbf{Blended Diffusion} & 81.42            & \textbf{29.13}  & \textbf{36.61}  & \textbf{19.16}  & \textbf{86.96} & \textbf{25.72}         & \textbf{23.56}          \\ 
\midrule
\textbf{Ours+P2P}                & \textbf{11.65}          & \underline{27.22}  & \underline{54.55}   & \underline{32.86}  & \underline{84.76} & \underline{25.02}         & \underline{22.10}          \\

\bottomrule
\end{tabular}}
\vspace{-0.2cm}
\caption{\textbf{Comparison of model-based editing results.}}
        \label{tab:end_to_end}
\end{table}

\vspace{-0.3cm}

\subsection{Influence of Guidance Scale}

\begin{table}[htbp]
\footnotesize
\centering
\setlength{\tabcolsep}{0.4mm}{
\begin{tabular}{c|c|c|cccc|cc}
\toprule
\multicolumn{2}{c|}{\textbf{Guidance Scale}}           & \textbf{Structure}          & \multicolumn{4}{c|}{\textbf{Background Preservation}} & \multicolumn{2}{c}{\textbf{CLIP Similariy}} \\ \midrule
\textbf{Inverse}          & \textbf{Forward}            & \textbf{Distance}$_{^{\times 10^3}}$ $\downarrow$ & \textbf{PSNR} $\uparrow$     & \textbf{LPIPS}$_{^{\times 10^3}}$ $\downarrow$  & \textbf{MSE}$_{^{\times 10^4}}$ $\downarrow$     & \textbf{SSIM}$_{^{\times 10^2}}$ $\uparrow$    & \textbf{Whole}  $\uparrow$          & \textbf{Edit}  $\uparrow$       \\ \midrule
0& 1                & 6.23 & 28.96 & 40.71 & 22.83 & 86.29 & 23.19 & 20.49 \\
0 & 2.5                & 8.03 & 28.27 & 44.59 & 25.90 & 85.89 & 23.89 & 21.09  \\
0 & 5                & 10.74 & 27.49 & 50.69 & 30.55 & 85.23 & 24.64 & 21.75  \\
0 & 7.5                & 13.38 & 26.86 & 56.66 & 35.29 & 84.56 & \underline{25.04} & 22.08  \\
1             & 1  &        \textbf{4.77} & \textbf{29.70} & \textbf{36.32} & \textbf{19.84} & \textbf{86.61} & 22.97 & 20.28   \\
1             & 2.5  &       \underline{5.74} & \underline{29.19} & \underline{39.57} & \underline{21.81} & \underline{86.35} & 23.87 & 21.01   \\
1             &  5   &         8.76 & 28.04 & 47.59 & 27.45 & 85.52 & 24.65 & 21.76   \\
1 & 7.5   & 11.65 & 27.22 & 54.54 & 32.86 & 84.76 & 25.02 & \underline{22.10} \\
2.5 & 1   & 20.17 & 25.87 & 67.73 & 49.33 & 83.44 & 21.28 & 19.02 \\
2.5 & 2.5   & 8.90 & 28.73 & 43.37 & 25.77 & 85.91 & 23.50 & 20.80 \\
2.5 & 5   & 9.01 & 28.27 & 47.05 & 27.77 & 85.55 & 24.62 & 21.70 \\
\rowcolor{lightyellow} 2.5 & 7.5   & 11.36 & 27.35 & 54.43 & 33.40 & 84.76 & \textbf{25.09} & \textbf{22.15} \\
5 & 1   & 70.65 & 19.85 & 155.46 & 203.41 & 75.92 & 16.38 & 15.84 \\
5 & 2.5   & 45.16 & 22.63 & 111.55 & 123.84 & 79.59 & 19.95 & 18.11 \\
5 & 5   & 28.85 & 25.18 & 79.92 & 70.91 & 82.35 & 22.60 & 20.09 \\
5 & 7.5   & 23.11 & 25.88 & 71.19 & 54.33 & 83.08 & 24.02 & 21.17 \\
7.5 & 1   & 97.45 & 17.62 & 203.49 & 309.26 & 71.92 & 14.38 & 14.79 \\
7.5 & 2.5   & 74.55 & 19.43 & 165.31 & 224.15 & 74.84 & 17.19 & 16.34 \\
7.5 & 5   & 56.22 & 21.53 & 131.04 & 154.97 & 77.57 & 19.82 & 18.07 \\
7.5 & 7.5   & 47.06 & 22.76 & 113.34 & 120.94 & 79.05 & 21.66 & 19.42 \\
\bottomrule
\end{tabular}}
\caption{\textbf{Ablation on the influence of guidance scale}}
\label{tab:ablation_guidance}
\end{table}

\begin{wrapfigure}{r}{0.6\textwidth}
    \vspace{-0.2cm}
\begin{center}
            \includegraphics[width=0.5\textwidth]{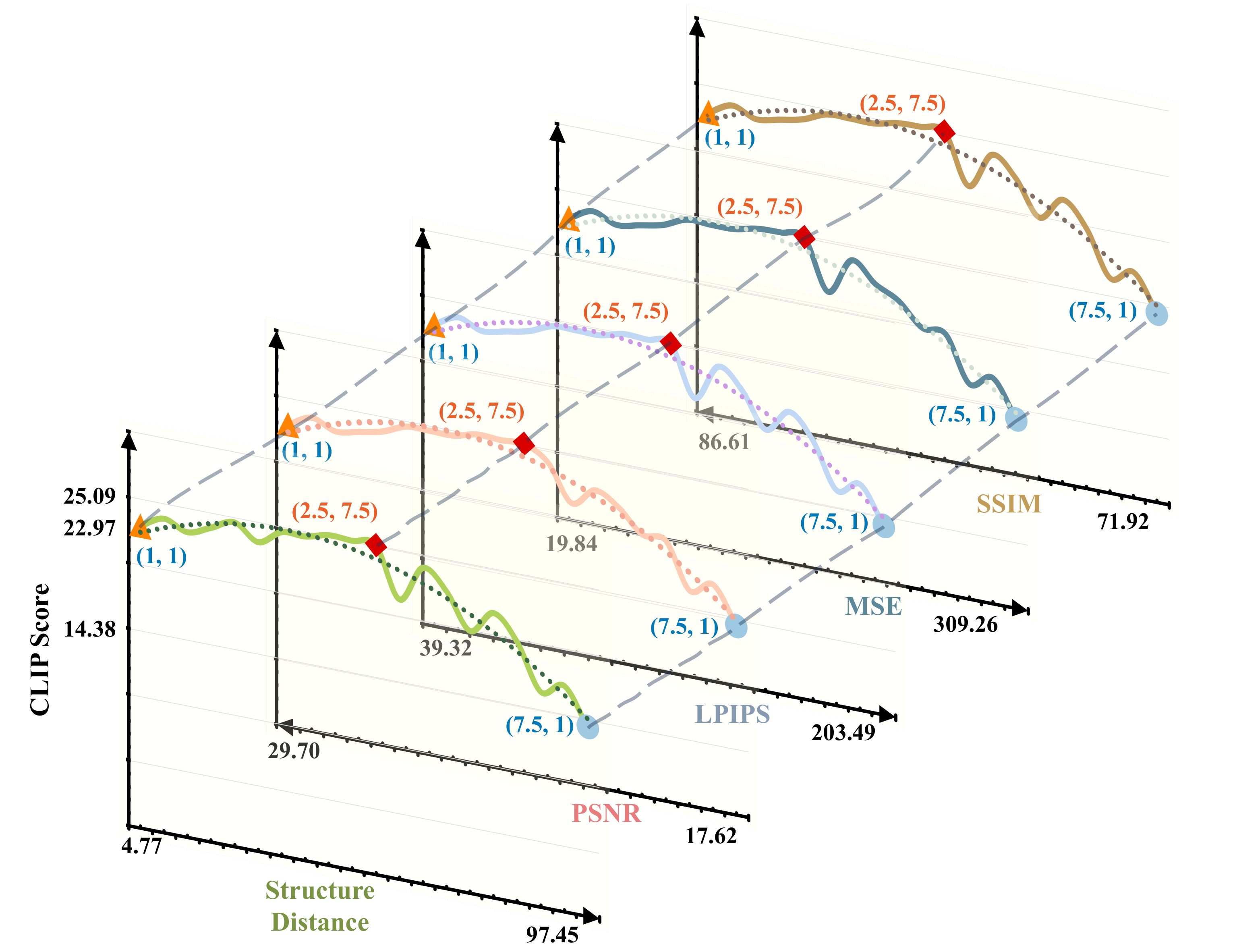}
        \end{center}
    \vspace{-0.3cm}
    \caption{\textbf{The impact of different inverse and forward guidance scales on evaluation metrics.} }
    \vspace{-0.3cm}
    \label{fig:ablation_guidance}
    \vspace{-0.2cm}
\end{wrapfigure}

In our experiments, we observed a significant impact of the guidance scale on the inversion and forward processes of DDIM, consequently affecting the editing results. Previous studies have typically employed an inversion guidance scale of 1/0, coupled with an empirical forward guidance scale of 7.5. However, no systematic experimental evidence determines the optimal combination of guidance scales for achieving the best editing performance, and elucidates how deviations in these guidance scales affect both reconstruction and editing. Hence, in this section, we present a comprehensive analysis of systematic results addressing this matter based on Table \ref{tab:ablation_guidance} and find that, in fact, an inverse guidance scale of $2.5$ and a forward guidance scale of $7.5$ reaches the best balance of essential content preservation and edit fidelity.

When keeping the inverse guidance scale constant, we observed that as the forward guidance scales increased gradually, there was an initial improvement in background preservation, followed by a decline. The inflection point was approximately at the inverse guidance scale being equal to the forward guidance scale (\emph{e.g.}, inverse with guidance scale 2.5, forward with guidance scale 2.5). In contrast, the CLIP similarity showed a consistently increasing trend.

Figure \ref{fig:ablation_guidance} enables a clear observation of a noticeable trade-off between essential content preservation and edit fidelity. The abscissa represents the sorted essential content preservation metrics, while the ordinate corresponds to the respective CLIP Similarity, aiming to illustrate the contrasting balance between these two major categories of metrics. Results show that an inverse guidance scale of 2.5 and a forward guidance scale of 7.5 show the best balance of editing and preservation.
The fundamental reason for this trade-off is that present editing methods lack the ability to accurately differentiate between regions that require modification and those that do not. This leads to an inherent conflict where successful and precise edits lead to substantial alterations of the source image while contradictory to the objective of essential content preservation.
This experimental observation emphasizes a distinct optimal range for the guidance scale on the evaluation metrics. A judiciously selected guidance scale can improve the alignment of the inverse and forward processes, and thus improve the editing performance.

\subsection{Ablation of Step and Interval}

To illustrate \OursMethod's performance in different diffusion steps and different add back intervals, we further provide results of \OursMethod added with Prompt-to-Prompt with step numbers of 20, 50, 100, and 500 in Table~\ref{tab:ablation_step}, and with an interval of 1, 2, 5, 10, 24, 49 in Table~\ref{tab:ablation_interval}. Results in Table~\ref{tab:ablation_step} show that \OursMethod is robust to different diffusion steps. Fewer steps will lead to a relatively better preservation of structure and background, bigger steps will have a better clip similarity since target text embedding brings more influence in the inference process. Table~\ref{tab:ablation_interval} shows the results of performing \OursMethod in interval steps, which leads to an update delay. Results show that with the interval increase, performance would become closer to DDIM Inversion with a larger structure/background distance. When the update is performed step-by-step, that is, when the interval is 1, \OursMethod performs best in terms of the overall metrics.

\begin{table}[htbp]
\footnotesize
\centering
\setlength{\tabcolsep}{0.4mm}{
\begin{tabular}{c|c|cccc|cc}
\toprule
\textbf{Metrics}        & \textbf{Structure}          & \multicolumn{4}{c|}{\textbf{Background Preservation}} & \multicolumn{2}{c}{\textbf{CLIP Similarity}} \\ \midrule
\textbf{Steps}            & \textbf{Distance}$_{^{\times 10^3}}$ $\downarrow$ & \textbf{PSNR} $\uparrow$     & \textbf{LPIPS}$_{^{\times 10^3}}$ $\downarrow$  & \textbf{MSE}$_{^{\times 10^4}}$ $\downarrow$     & \textbf{SSIM}$_{^{\times 10^2}}$ $\uparrow$    & \textbf{Whole}  $\uparrow$          & \textbf{Edited}  $\uparrow$       \\ \midrule
\textbf{20} & \textbf{10.60} & \textbf{27.49} & \textbf{51.80} & \textbf{30.62} & \textbf{84.94} & 24.73 & 21.75 \\ 
\textbf{50}                & 11.65 & 27.22 & 54.55 & 32.86 & 84.76 & 25.02 & 22.10        \\
\textbf{100} & 12.22 & 27.01 & 56.00 & 34.49 & 84.53 & 25.18 & 22.28 \\ 
\textbf{500} & 13.00 & 26.82 & 57.44 & 35.76 & 84.39 & \textbf{25.30} & \textbf{22.42} \\ 
\bottomrule
\end{tabular}}
\caption{\textbf{Ablation of different inference steps.}}
\label{tab:ablation_step}
\end{table}

\begin{table}[htbp]
\footnotesize
\centering
\setlength{\tabcolsep}{0.4mm}{
\begin{tabular}{c|c|cccc|cc}
\toprule
\textbf{Metrics}        & \textbf{Structure}          & \multicolumn{4}{c|}{\textbf{Background Preservation}} & \multicolumn{2}{c}{\textbf{CLIP Similarity}} \\ \midrule
\textbf{Interval}            & \textbf{Distance}$_{^{\times 10^3}}$ $\downarrow$ & \textbf{PSNR} $\uparrow$     & \textbf{LPIPS}$_{^{\times 10^3}}$ $\downarrow$  & \textbf{MSE}$_{^{\times 10^4}}$ $\downarrow$     & \textbf{SSIM}$_{^{\times 10^2}}$ $\uparrow$    & \textbf{Whole}  $\uparrow$          & \textbf{Edited}  $\uparrow$       \\ \midrule
\textbf{1} & \textbf{11.65} & \textbf{27.22} & 54.55 & \textbf{32.86} & \textbf{84.76} & 25.02 & 22.10        \\
\textbf{2} & 11.83 & 27.15 & \textbf{54.53} & 33.32 & 84.67 & \textbf{25.06} & 22.11 \\ 
\textbf{5} & 13.06 & 26.86 & 57.30 & 35.32 & 84.43 & 25.05 & 22.11 \\ 
\textbf{10} & 16.18 & 26.03 & 66.37 & 42.24 & 83.60 & 24.98 & 22.13 \\ 
\textbf{24} & 24.08 & 24.20 & 77.73 & 66.99 & 82.41 & 24.91 & 22.05 \\ 
\textbf{49} & 47.05 & 21.21 & 128.64 & 126.34 & 78.19 & 24.81 & \textbf{22.22} \\ 
\bottomrule
\end{tabular}}
\caption{\textbf{Ablation of performing \OursMethod in interval steps.}}
\label{tab:ablation_interval}
\end{table}

\subsection{Results of Different Editing Types}

We provide the performance of \OursMethod added to Prompt-to-Prompt~\citep{hertz2022prompt} in Table~\ref{tab:ablation_type}. The results vary across different editing types, with type 0 performing quite closely in line with its performance across all categories, as it involves random volunteer-selected images and editorial instructions. In the editing types that Prompt2Prompt struggles with, such as adding objects (type 2), deleting objects (type 3), and modifying object pose (type 5), the model shows minimal changes, resulting in a relatively better evaluation result in essential content preservation metrics. Anomaly, type 5 shows the highest Clip Similarity on editing object poses. We infer the reason lies in the insensitivity of the CLIP model on the object pose and leads to a similarity in features of the source and target prompt. And since the source prompt is written by Blip2, which has a high CLIP similarity to the source image, the images with minor alterations in type 5 tend to have a better CLIP Similarity to the source prompt. Type 8 (change background) and 9 (change style) have a bad Structure Distance because the areas that need modification are relatively large. For type 9, the whole image is required for editing. Thus, we do not report background preservation metric and Whole Image CLIP Similarity, which is the same as Edit Region Clip Similarity.

\begin{table}[htbp]
\footnotesize
\centering
\setlength{\tabcolsep}{0.4mm}{
\begin{tabular}{c|c|cccc|cc}
\toprule
\textbf{Metrics}        & \textbf{Structure}          & \multicolumn{4}{c|}{\textbf{Background Preservation}} & \multicolumn{2}{c}{\textbf{CLIP Similarity}} \\ \midrule
\textbf{Type}            & \textbf{Distance}$_{^{\times 10^3}}$ $\downarrow$ & \textbf{PSNR} $\uparrow$     & \textbf{LPIPS}$_{^{\times 10^3}}$ $\downarrow$  & \textbf{MSE}$_{^{\times 10^4}}$ $\downarrow$     & \textbf{SSIM}$_{^{\times 10^2}}$ $\uparrow$    & \textbf{Whole}  $\uparrow$          & \textbf{Edited}  $\uparrow$       \\ \midrule
\textbf{0} & 11.73 & 28.58 & 50.07 & 30.18 & 85.87 & 25.09 & 22.66       \\
\textbf{1} &  12.60 & 26.74 & 55.76 & 29.75 & 84.34 & 24.48 & 20.11      \\
\textbf{2} &  10.09 & 27.55 & 53.68 & 30.71 & 86.72 & 25.12 & 23.20      \\
\textbf{3} &  \textbf{9.65} & 23.97 & 78.78 & 53.06 & 78.37 & 23.76 & 17.56      \\
\textbf{4} &  12.59 & 28.12 & 51.97 & 23.13 & 85.85 & 24.85 & 22.58      \\
\textbf{5} & 10.08 & 26.57 & 60.23 & 33.42 & 82.98 & \textbf{26.24} & 22.38       \\
\textbf{6} &  10.53 & 26.40 & 55.34 & 35.91 & 83.54 & 25.55 & 20.95      \\
\textbf{7} &  11.21 & \textbf{30.39} & 40.38 & \textbf{17.47} & \textbf{89.14} & 25.81 & 23.87      \\
\textbf{8} & 12.23 & 27.40 & \textbf{39.03} & 30.70 & 87.68 & 24.34 & 22.01       \\
\textbf{9} & 14.50 & - & - & - & - & - & \textbf{26.04}       \\
\bottomrule
\end{tabular}}
\caption{\textbf{Performance of \OursMethod in different editing types.}}
\label{tab:ablation_type}
\end{table}

\section{Qualitative Results}
\label{supp:qualitative_results}

\vspace{-0.2cm}

Due to the page limit, we do not provide lots of visualization results in the main paper. In this section, we provide a comparison of visualization for further verification of quantitative results. 

\vspace{-0.2cm}

\begin{figure}[htbp]
        \begin{center}
            \includegraphics[width=0.75\textwidth]{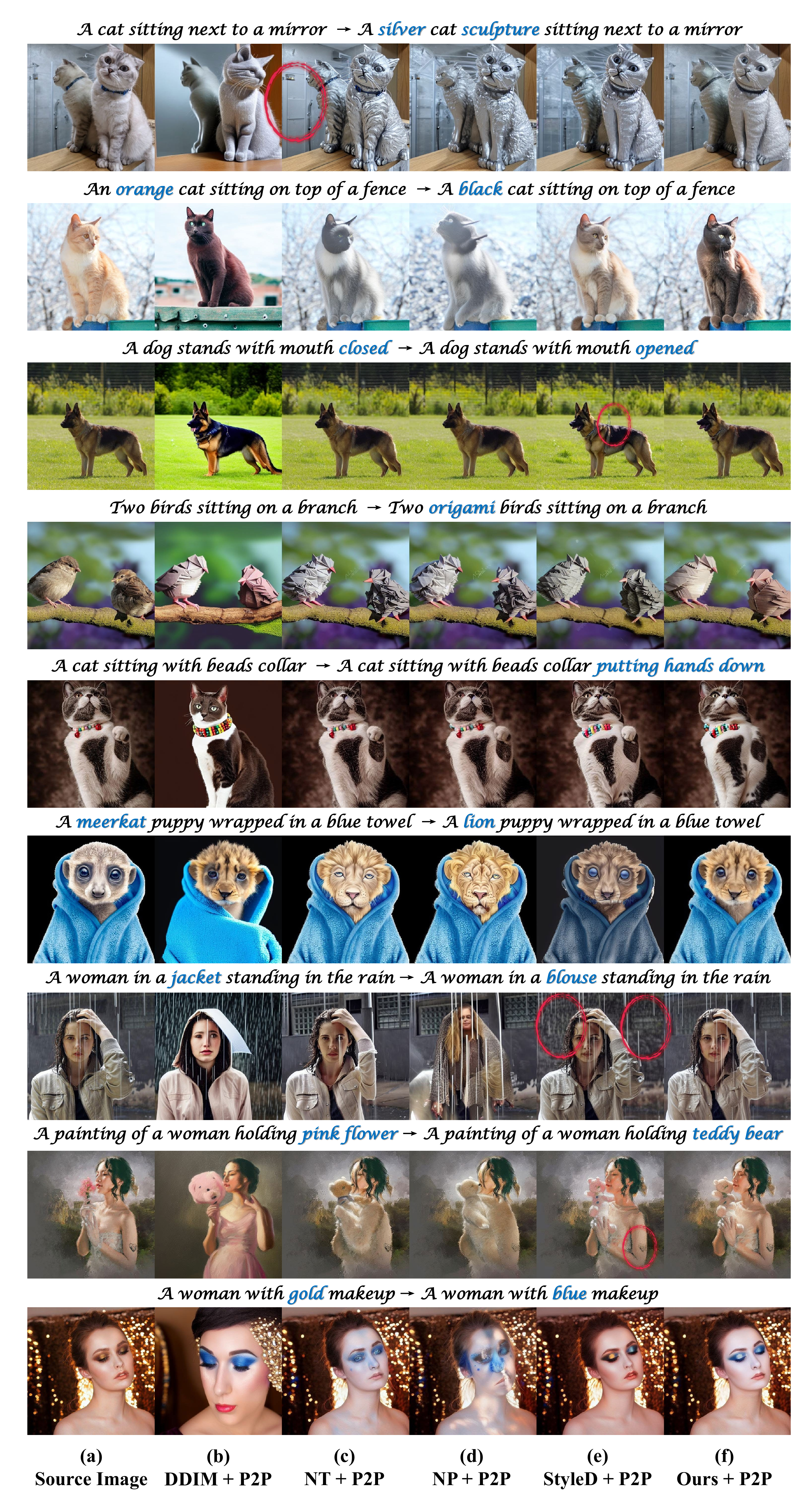}
        \end{center}
    \caption{\textbf{Visualization results of different inversion methods combined with Prompt-to-Prompt (P2P)~\citep{hertz2022prompt}.} The source image is shown in col (a). We compare (f) \OursMethod with different inversion techniques: (b) DDIM Inversion (DDIM)~\citep{song2020denoising}, (c) Null-Text Inversion (NT)~\citep{mokady2023null}, (d) Negative-Prompt Inversion (NP)~\citep{miyake2023negative}, and (e) StyleDiffusion (StyleD)~\citep{li2023stylediffusion}. The source and target prompt are shown at the top of each row. The improvements are mostly tangible, and we circle some of the subtle discrepancies w/o \OursMethod in red.}
        \label{fig:supp_p2p}
\end{figure}

\begin{figure}[htbp]
        \begin{center}
            \includegraphics[width=0.75\textwidth]{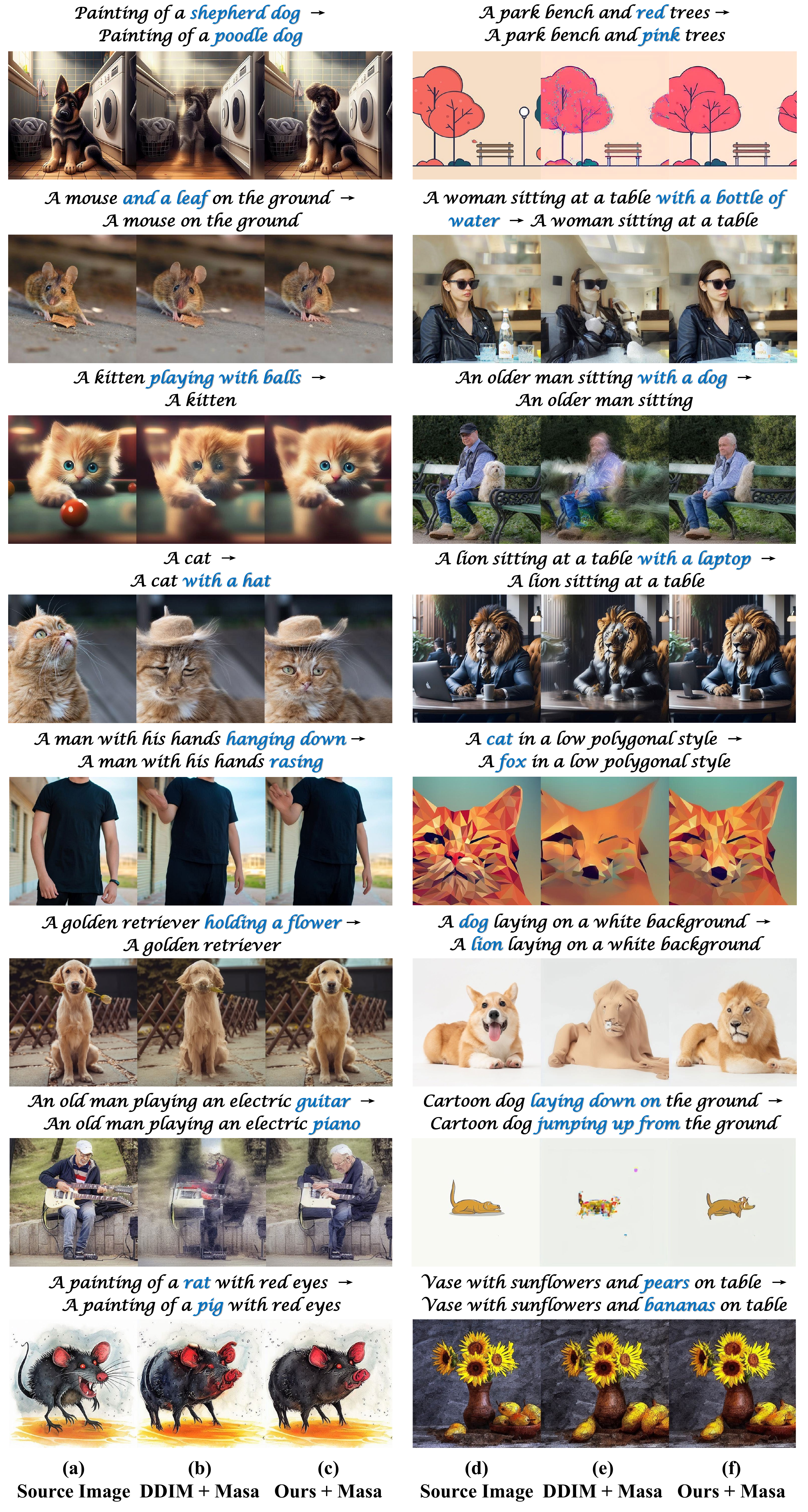}
        \end{center}
    \caption{\textbf{Visualization results of MasaCtrl (Masa)~\citep{cao2023masactrl} w/ and w/o \OursMethod.} The source image is shown in col (a) and (d). The col (b) and (e) show results w/o \OursMethod. The col (c) and (f) show results w/ \OursMethod. The source and target prompt are shown at the top of each row. }
        \label{fig:supp_masactrl}
\end{figure}

\begin{figure}[htbp]
        \begin{center}
            \includegraphics[width=0.75\textwidth]{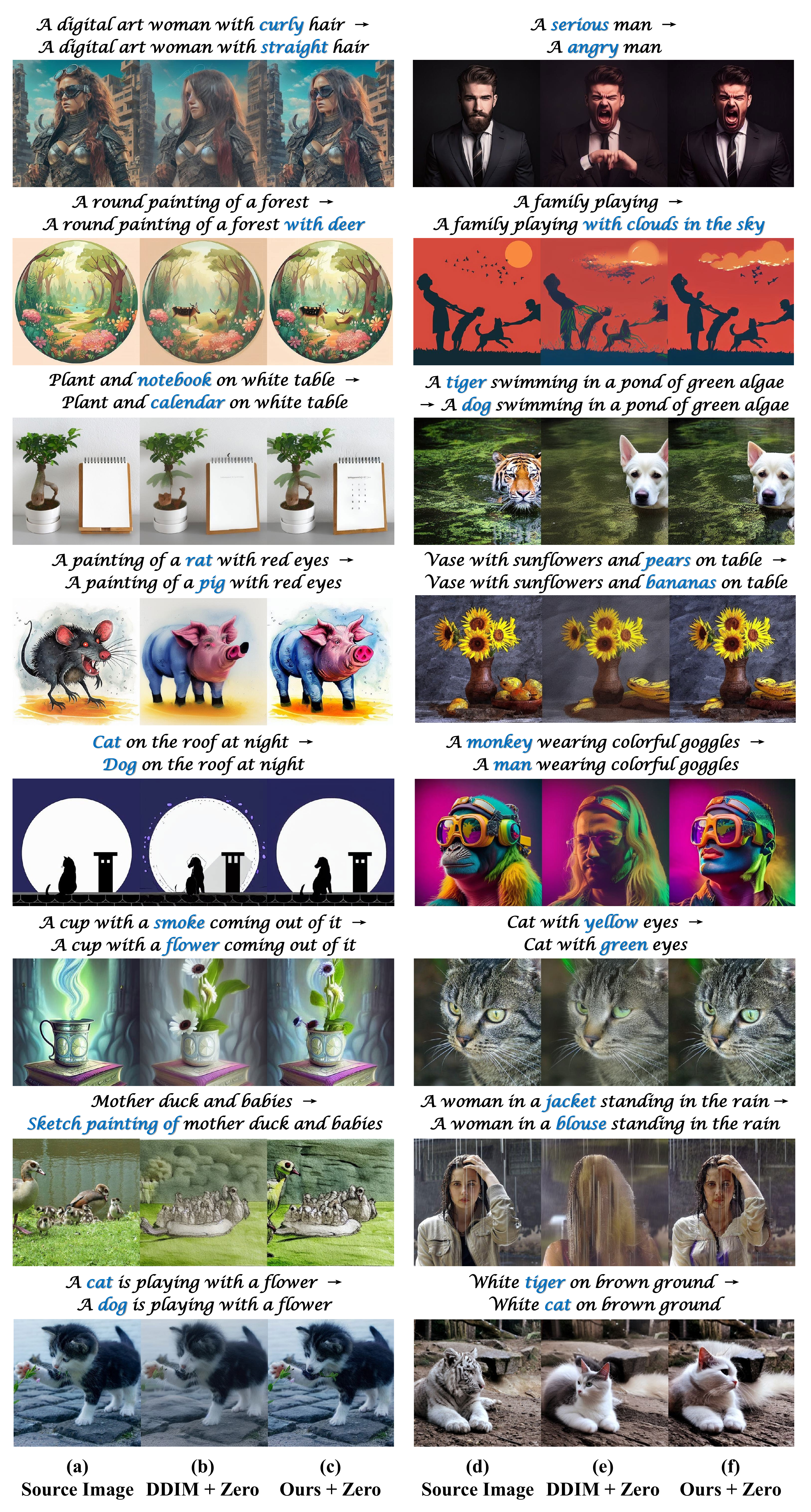}
        \end{center}
    \caption{\textbf{Visualization results of Pix2Pix-Zero (Zero)~\citep{parmar2023zero} w/ and w/o \OursMethod.} The source image is shown in col (a) and (d). The col (b) and (e) show results w/o \OursMethod. The col (c) and (f) show results w/ \OursMethod. The source and target prompt are shown at the top of each row. }
        \label{fig:supp_zero}
\end{figure}

\begin{figure}[htbp]
        \begin{center}
            \includegraphics[width=0.75\textwidth]{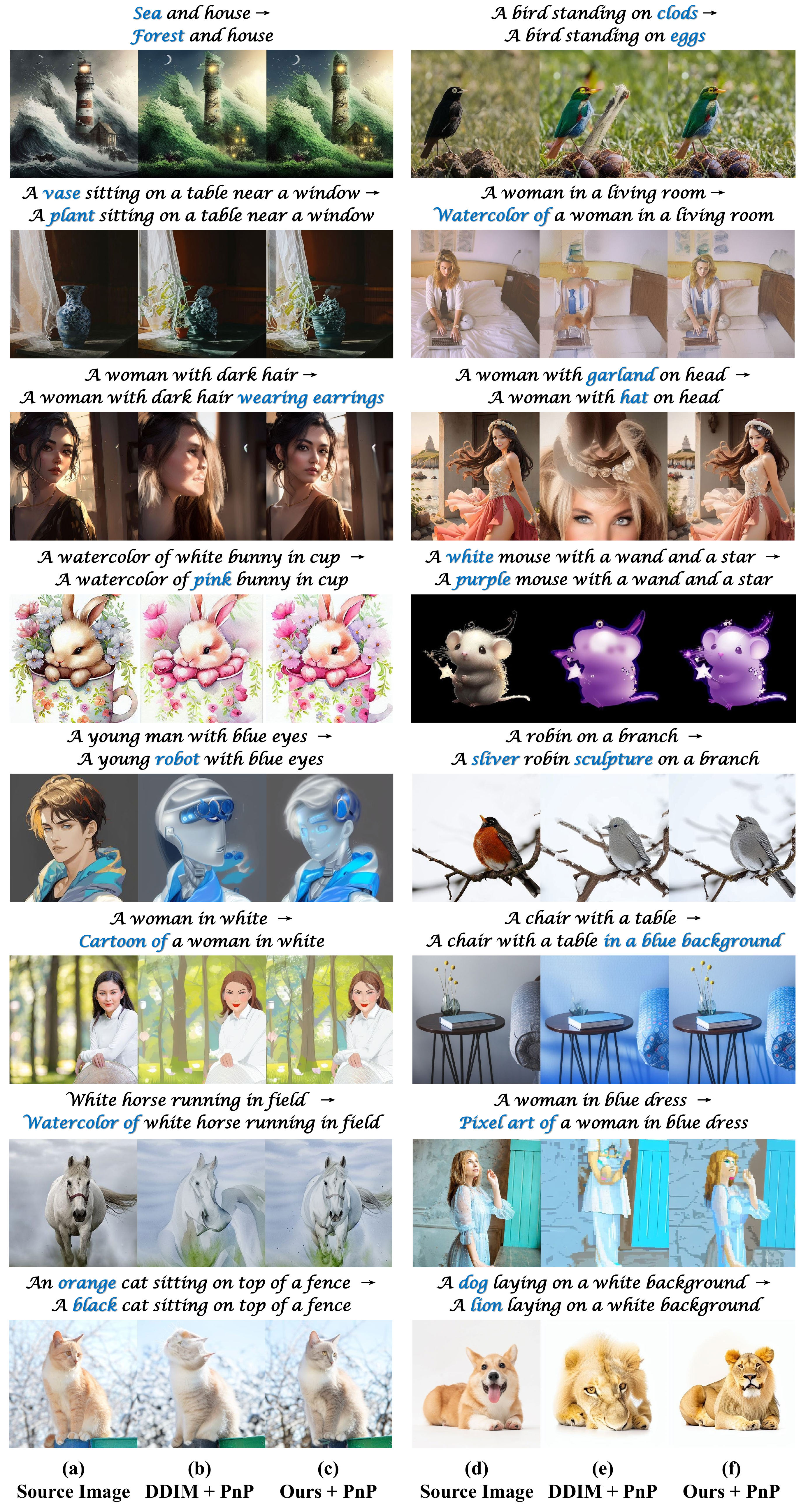}
        \end{center}
    \caption{\textbf{Visualization results of Plug-and-Play (PnP)~\citep{tumanyan2023plug} w/ and w/o \OursMethod.} The source image is shown in col (a) and (d). The col (b) and (e) show results w/o \OursMethod. The col (c) and (f) show results w/ \OursMethod. The source and target prompt are shown at the top of each row. }
        \label{fig:supp_pnp}
\end{figure}

\paragraph{Comparison with different inversion-based editing.} Figure~\ref{fig:supp_p2p} shows the comparison of different inversion methods combined with Prompt-to-Prompt~\citep{hertz2022prompt}. Figure~\ref{fig:supp_masactrl}, Figure~\ref{fig:supp_zero}, and Figure~\ref{fig:supp_pnp} shows the visualization results of MasaCtrl~\citep{cao2023masactrl}, Pix2Pix-Zero~\citep{parmar2023zero}, and Plug-and-Play~\citep{tumanyan2023plug} w/ and w/o \OursMethod.

\vspace{-0.2cm}

\begin{figure}[htbp]
        \begin{center}
            \includegraphics[width=0.75\textwidth]{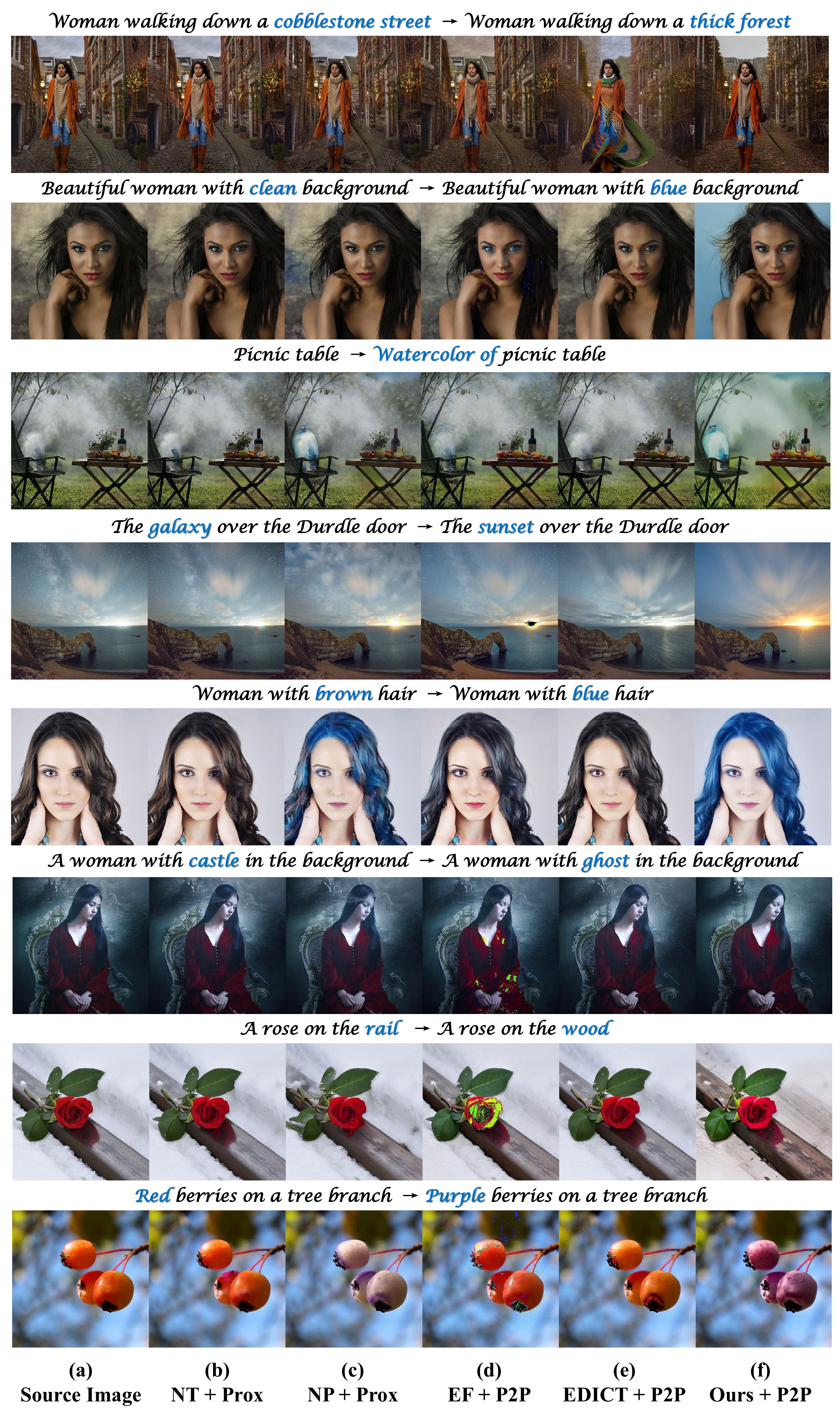}
        \end{center}
    \caption{\textbf{Visualization results of different essential content preservation methods.} The source image is shown in col (a). We compare (f) \OursMethod added to Prompt-to-Prompt (P2P)~\citep{hertz2022prompt} with different combined methods: (b) Null-Text Inversion (NT)~\citep{mokady2023null} added to Proximal Guidance (Prox)~\citep{han2023improving}, (c) Negative-Prompt Inversion (NP)~\citep{miyake2023negative} added to Proximal Guidance, (d) Edit Friendly DDPM (EF)~\citep{huberman2023edit} added to Prompt-to-Prompt, and (e) EDICT~\citep{wallace2023edict} added to Prompt-to-Prompt. The source and target prompt are shown at the top of each row.}
        \label{fig:supp_back}
\end{figure}

\paragraph{Comparison with essential content preservation methods.} Visualization results of essential content preservation methods are shown in Figure~\ref{fig:supp_back}, including Proximal Guidance~\citep{han2023improving}, Edit Friendly DDPM~\citep{huberman2023edit}, and EDICT~\citep{wallace2023edict}.

\vspace{-0.2cm}

\begin{figure}[htbp]
        \begin{center}
            \includegraphics[width=0.75\textwidth]{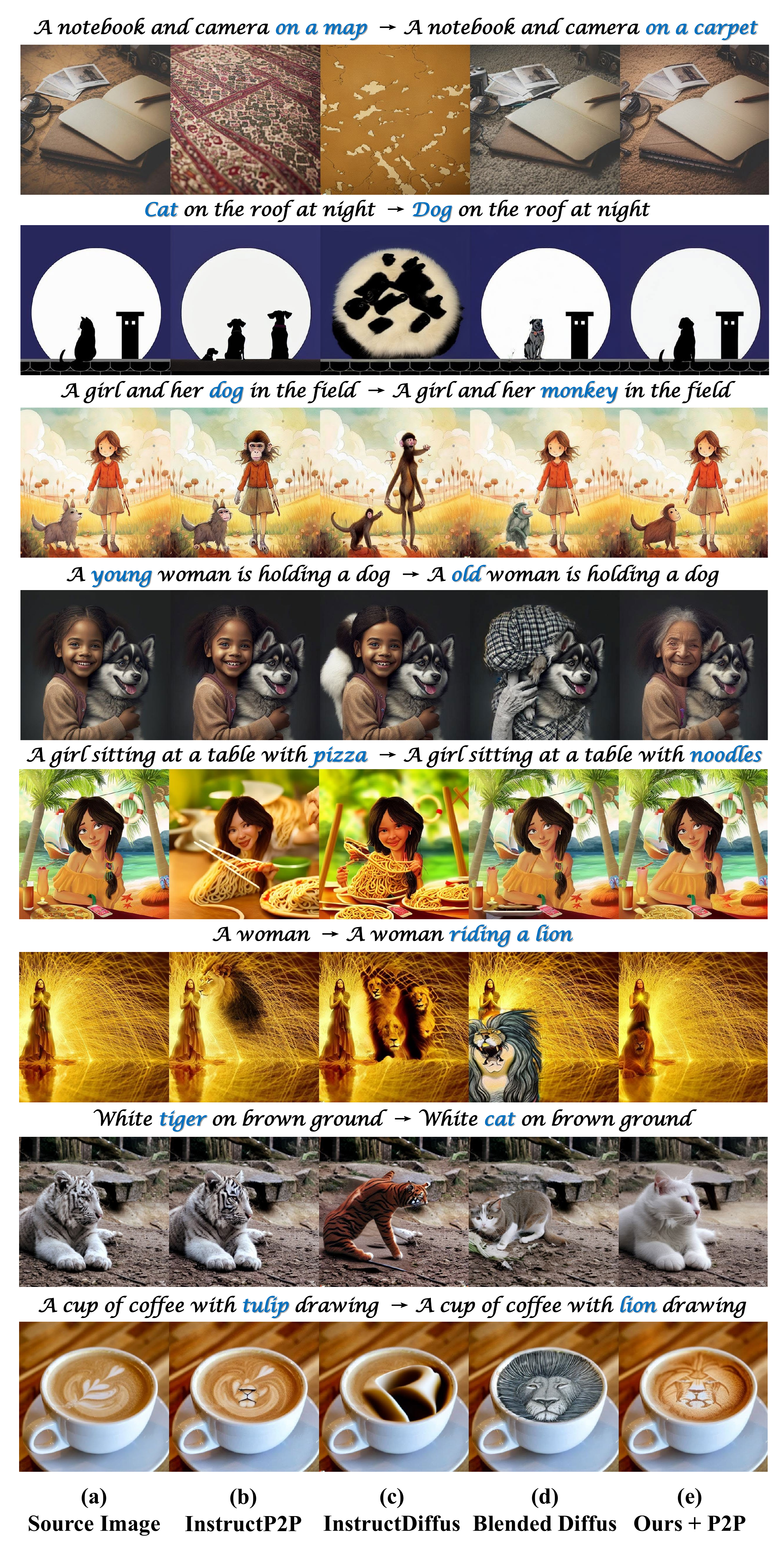}
        \end{center}
    \caption{\textbf{Visualization results of different essential content preservation methods.} The source image is shown in col (a). We compare (f) \OursMethod added to Prompt-to-Prompt (P2P)~\citep{hertz2022prompt} with different model-based editing methods: (b) InstructPix2Pix (InstructP2P)~\citep{brooks2023instructpix2pix}, (c) InstructDiffusion (InstructDiffus)~\citep{geng2023instructdiffusion}, and (d) Blended Latent Diffusion (Blended Diffus)~\citep{avrahami2023blended}. The source and target prompt are shown at the top of each row.}
        \label{fig:supp_model}
\end{figure}

\paragraph{Comparison with model-based editing.} Visualization results of model-based editing are shown in Figure~\ref{fig:supp_model}, including InstructPix2Pix~\citep{brooks2023instructpix2pix}, InstructDiffusion~\citep{geng2023instructdiffusion}, and Blended Latent Diffusion~\citep{avrahami2023blended}.

\section{Limitations and Future Works}
\label{supp:limitation}

\begin{figure}[htbp]
        \begin{center}
            \includegraphics[width=0.75\textwidth]{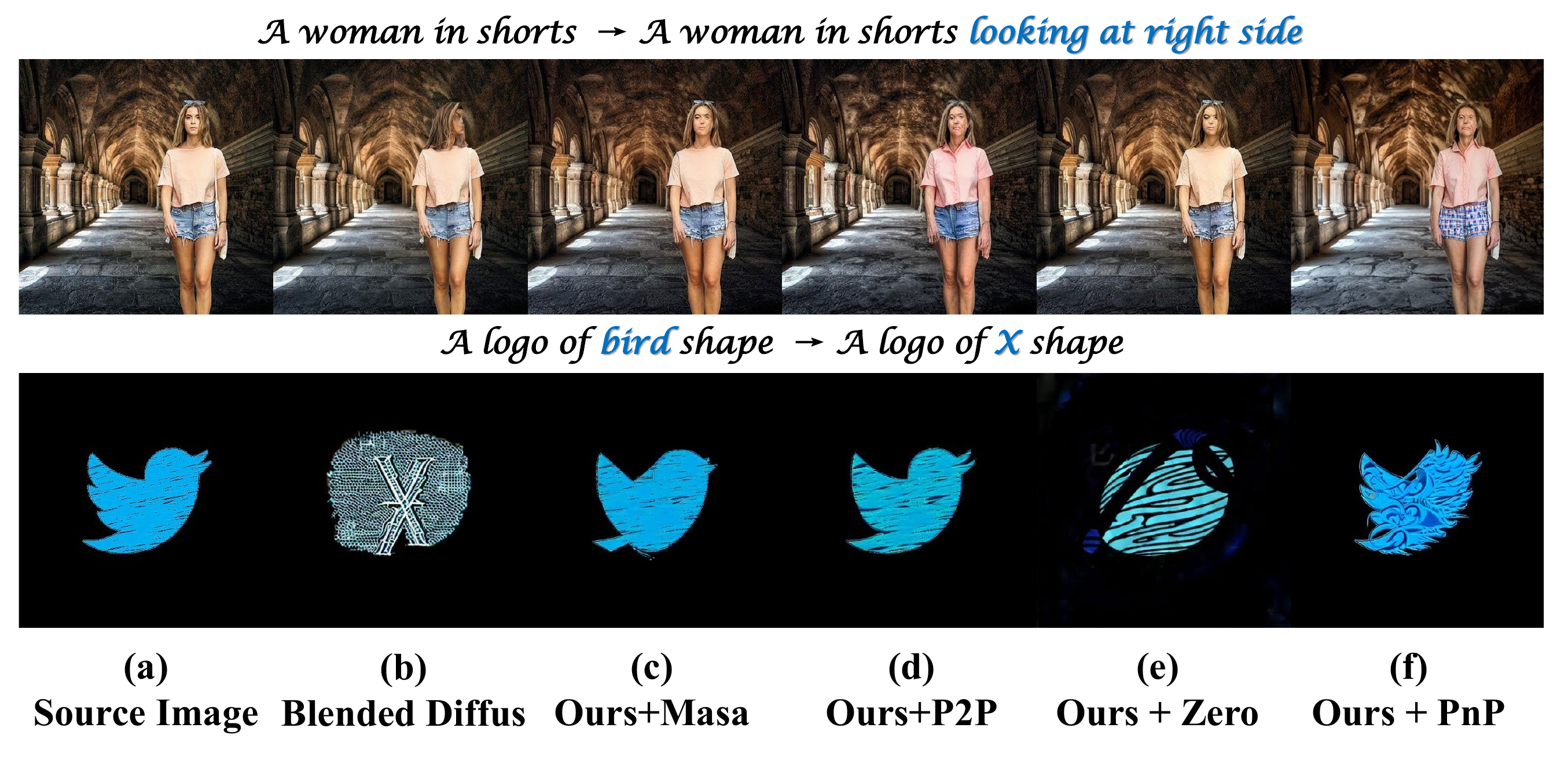}
        \end{center}
    \caption{\textbf{Visualization results of failure cases in existing diffusion-based editing methods.} (a) source image; (b) Blended Latent Diffusion (Blended Diffus)~\citep{avrahami2023blended}; (c) \OursMethod added to MasaCtrl (Masa)~\citep{cao2023masactrl}; (d) \OursMethod added to Prompt-to-Prompt (P2P)~\citep{hertz2022prompt}; (e) \OursMethod added to Pix2Pix-Zero (Zero)~\citep{parmar2023zero}; (f)\OursMethod added to Plug-and-Play (PnP)~\citep{tumanyan2023plug}. The source and target prompt are shown at the top of each row.}
        \label{fig:supp_fail}
\end{figure}

Since the performance of \OursMethod is strongly connected to existing diffusion-based editing methods, our method inherits most of their limitations. Although \OursMethod boosts existing editing techniques' performance, it is still unable to bring about fundamental changes in the editing model performance, which is unstable, with a low success rate. In Figure~\ref{fig:supp_fail}, we have chosen specific cases in which Blended Latent Diffusion, along with the ground truth mask, succeeds, whereas all other editing methods fail. This demonstrates the inherent capability of diffusion models to perform corresponding edits. However, existing diffusion-based editing algorithms lack the capability of realization without giving explicit masks. Moreover, although \OursMethod leads to a better performance on average, success is not guaranteed in every case.

\OursMethod may also lead to ethical issues that are worthy of consideration. The data used in the training of diffusion models unavoidably contain personally identifiable information, social biases, and violent content, which will also influence the editing results of our model. \OursMethod can be misused or modified to produce contradictory results and lead to potential negative societal impacts (\emph{e.g.}, arbitrary modification on private photo). We believe these issues should be considered, and we need to design and engineer AI capabilities to fulfill their intended functions while possessing the ability to detect and avoid unintended consequences and unintended behavior.

We hope that this work can motivate future research with a focus on diffusion-based editing for higher essential content preservation and edit fidelity.  Specifically, future directions include but are not limited to (1) an extension to video editing, (2) editing models with higher success rates and more editable scenes, and (3) a more comprehensive metric evaluation system to evaluate the effectiveness of editing.

\end{document}